%% file: paper.tex
\documentclass[runningheads]{llncs}
\usepackage{graphicx}
\usepackage{amsmath,amssymb}

\input{preamble}

\begin{document}

\title{On Learning Associations of Faces and Voices}
\titlerunning{On Learning Associations of Faces and Voices}


\author{%
	Changil Kim\inst{1} \and
	Hijung Valentina Shin\inst{2} \and
	Tae-Hyun Oh\inst{1} \and
	Alexandre Kaspar\inst{1} \and
	Mohamed Elgharib\inst{3} \and
	Wojciech Matusik\inst{1}}
\authorrunning{C.\ Kim et al.}


\institute{%
	MIT CSAIL \and
	Adobe Research \and
	QCRI}

\maketitle


\setcounter{footnote}{0}

\input{abstract}

\input{figures}

\input{intro}

\input{related}

\input{human}

\input{machine}

\input{conclusion}

\input{acknowledgments}


\newcommand{\suppmark}{A.}
\renewcommand\thefigure{\suppmark\arabic{figure}}
\renewcommand\thetable{\suppmark\arabic{table}}
\renewcommand\thesection{\suppmark\arabic{section}}
\renewcommand\theequation{\suppmark\arabic{equation}}
\setcounter{figure}{0}
\setcounter{table}{0}
\setcounter{section}{0}
\setcounter{equation}{0}

\input{supp_content}


\bibliographystyle{splncs04}
\bibliography{paper}

\end{document}

%% file: abstract.tex
\begin{abstract}
  In this paper, we study the associations between human faces and voices. Audiovisual integration, specifically the integration of facial and vocal information is a well-researched area in neuroscience. It is shown that the overlapping information between the two modalities plays a significant role in perceptual tasks such as speaker identification. Through an online study on a new dataset we created, we confirm previous findings that people can associate unseen faces with corresponding voices and vice versa with greater than chance accuracy. We computationally model the overlapping information between faces and voices and show that the learned cross-modal representation contains enough information to identify matching faces and voices with performance similar to that of humans. Our representation exhibits correlations to certain demographic attributes and features obtained from either visual or aural modality alone. We release our dataset of audiovisual recordings and demographic annotations of people reading out short text used in our studies.
  \keywords{face-voice association \and multi-modal representation learning}
\end{abstract}

%% file: figures.tex
\newcommand{\addHumanPerformance}{%
\begin{table}[t]
  \centering
  \caption{The average performance of Amazon Mechanical Turk participants in each of the four experimental conditions.}
  \vspace{-.1cm}
  \scriptsize
  \begin{tabular}{@{}%
      C{0.40\linewidth}%
      L{0.10\linewidth}%
      L{0.10\linewidth}%
      L{0.15\linewidth}%
      L{0.15\linewidth}%
      @{}}
    \toprule
    Demographic constraints & Mean & SD & $t$ (n) & $p$-value\\
    \midrule
    G & 71.4\% & 13.6\% & 13.17 (70) & $p < 0.001$\\
    G/E & 65.0\% & 13.0\% & 9.65 (70) &  $p < 0.001$\\
    G/E/F/A& 58.4\%& 13.8\%& 5.20 (73)& $p < 0.001$\\
    G/E/F/A, F $\rightarrow$ V & 55.2\% & 12.2\% & 3.69 (75) & $p < 0.001$\\
    \bottomrule
  \end{tabular}
  \label{tab:human_performance}
  \vspace{-.4cm}
\end{table}
}

\newcommand{\addModelPerformanceEmbed}{%
\begin{wrapfigure}{r}{0.4\linewidth}
  \includegraphics[width=\linewidth]{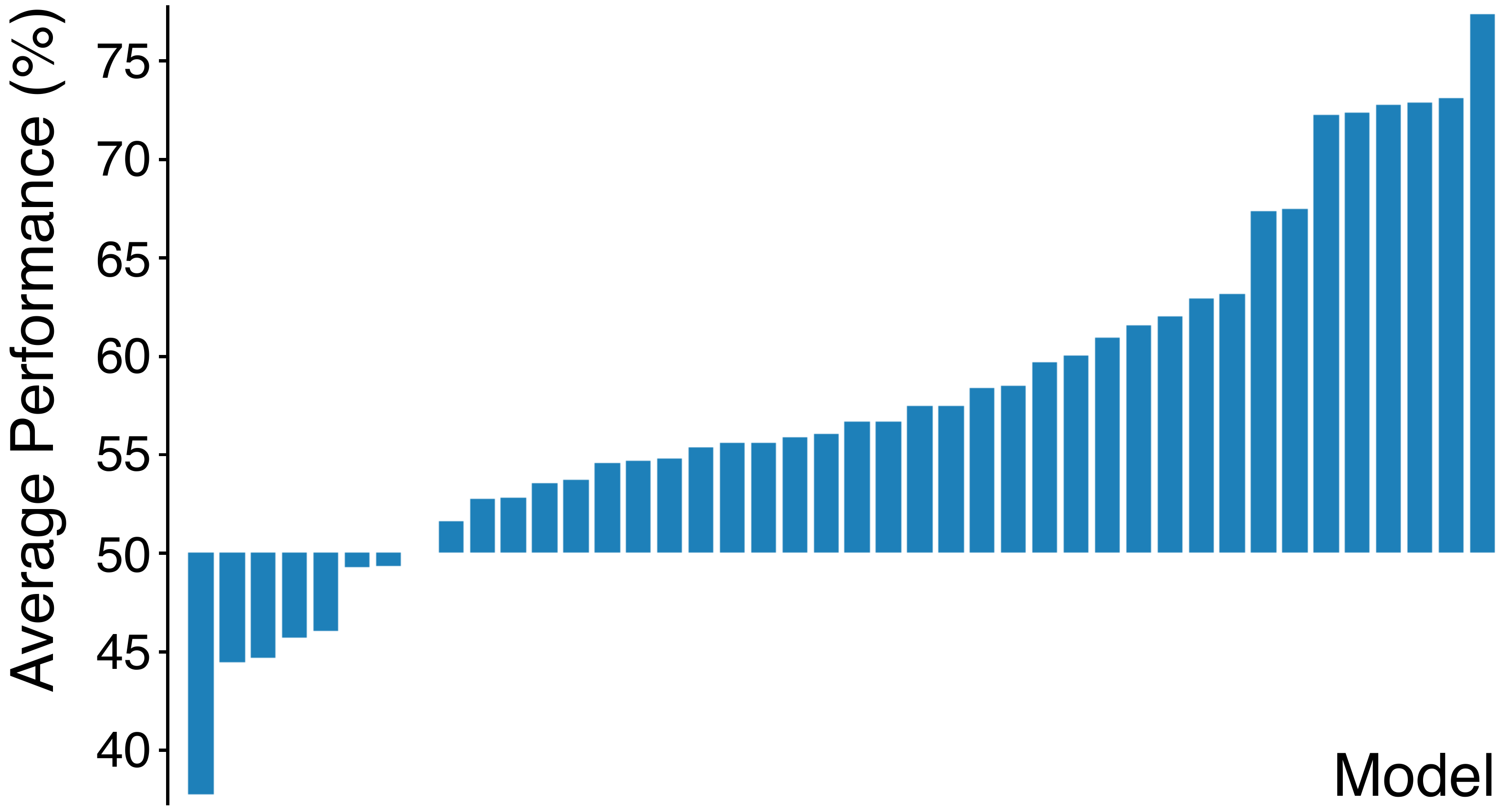}
\end{wrapfigure}
}

\newcommand{\addDemography}{%
\begin{table}[t]
  \centering
  \caption{The demographic distributions of our dataset for user studies and the VoxCeleb~\cite{Nagrani2017} test set. The fluency denotes whether the English language is the speaker's first language (Y) or not (N). The ethnicity denotes one of the following groups: (1)~American Indian; (2)~Asian and Pacific Islander; (3)~black or African American; (4)~Hispanic or Latino; (5)~non-Hispanic white; (6)~others.}
  \vspace{-.1cm}
  \scriptsize
  \begin{tabular}{@{}%
      L{0.1\linewidth}%
      R{0.04\linewidth}%
      R{0.04\linewidth}%
      R{0.04\linewidth}%
      R{0.04\linewidth}%
      R{0.04\linewidth}%
      R{0.04\linewidth}%
      R{0.04\linewidth}%
      R{0.04\linewidth}%
      R{0.05\linewidth}%
      R{0.04\linewidth}%
      R{0.05\linewidth}%
      R{0.04\linewidth}%
      R{0.04\linewidth}%
      R{0.04\linewidth}%
      R{0.04\linewidth}%
      R{0.04\linewidth}%
      R{0.04\linewidth}%
      R{0.05\linewidth}%
      @{}}
    \toprule
    \multirow{2}{*}[-.5mm]{Dataset} & \multicolumn{2}{c}{Gender} & \multicolumn{6}{c}{Ethnicity} & \multicolumn{2}{c}{Fluency} & \multicolumn{8}{c}{Age group} \\
    \cmidrule(l{3pt}r{1pt}){2-3} \cmidrule(l{3pt}r{1pt}){4-9} \cmidrule(l{3pt}r{1pt}){10-11} \cmidrule(l{3pt}r{1pt}){12-19}
             &  m. &  f. &   1 &   2 &   3 &   4 &   5 &   6 &   Y &   N & $\le$19 & 20s & 30s & 40s & 50s & 60s & 70s & $\ge$80 \\
    \midrule
    Ours     &  \textbf{95} &  86 &   5 &  30 &  14 &  15 &  \textbf{97} &  20 & \textbf{134} &  47 &       6 & \textbf{101} &  53 &  14 &   4 &   3 &   0 &       0 \\
    VoxCeleb & \textbf{150} & 100 &   1 &  10 &  19 &  13 & \textbf{189} &  18 & \textbf{223} &  27 &       2 &  27 &  \textbf{77} &  58 &  43 &  21 &  14 &       8 \\
    \bottomrule
  \end{tabular}
  \label{tab:demography}
  \vspace{-.4cm}
\end{table}
}

\newcommand{\addMachinePerformance}{%
\begin{table}[t]
  \centering
  \caption{The performance of our model measured on the VoxCeleb test set. Experiments are controlled with varying demographic grouping: without control~(--), within the same gender~(G), ethnic group~(E), both of the two~(G/E), and on the largest and most homogeneous group, i.e., non-Hispanic white, male native speakers in their 30s.}
  \vspace{-.2cm}
  \scriptsize
  \begin{tabular}{@{}%
      L{0.13\linewidth}%
      L{0.03\linewidth}%
      L{0.13\linewidth}%
      L{0.13\linewidth}%
      L{0.13\linewidth}%
      L{0.13\linewidth}%
      L{0.13\linewidth}%
      @{}}
    \toprule
    \multirow{2}{*}[-.5mm]{Direction}  && \multicolumn{5}{c}{Demographic grouping of test samples} \\
    \cmidrule{3-7}
                                && -- & G & E & G/E & G/E/F/A \\
    \midrule
    V $\rightarrow$ F           && 78.2\% & 62.9\% & 76.4\% & 61.6\% & 59.0\% \\
    F $\rightarrow$ V           && 78.6\% & 61.6\% & 76.7\% & 61.2\% & 56.8\% \\
    \bottomrule
  \end{tabular}
  \label{tab:machine_performance}
  \vspace{-.4cm}
\end{table}
}

\newcommand{\addTSNE}{%
\begin{figure}[t]
  \centering
  \includegraphics[width=0.25\linewidth]{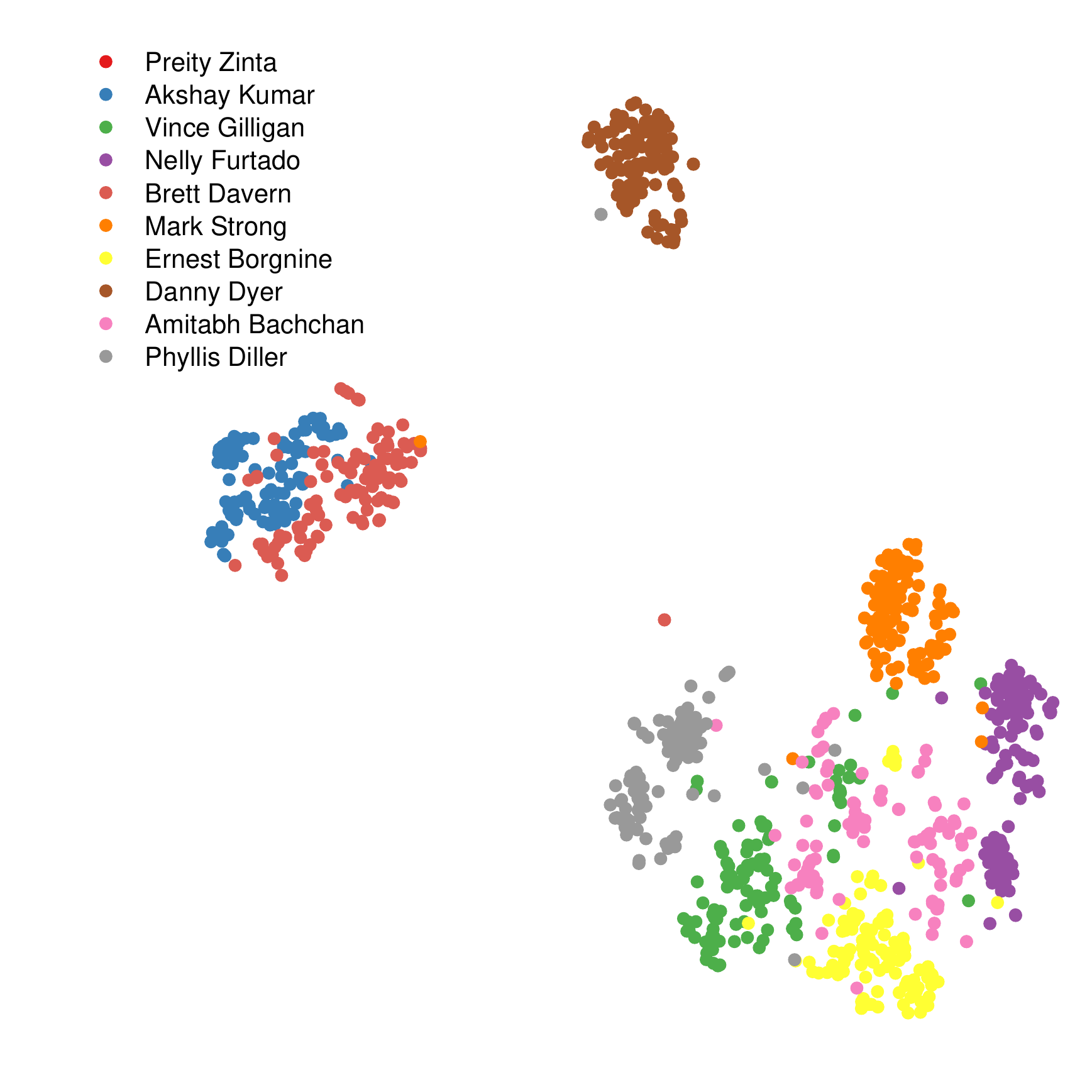}\quad%
  \includegraphics[width=0.25\linewidth]{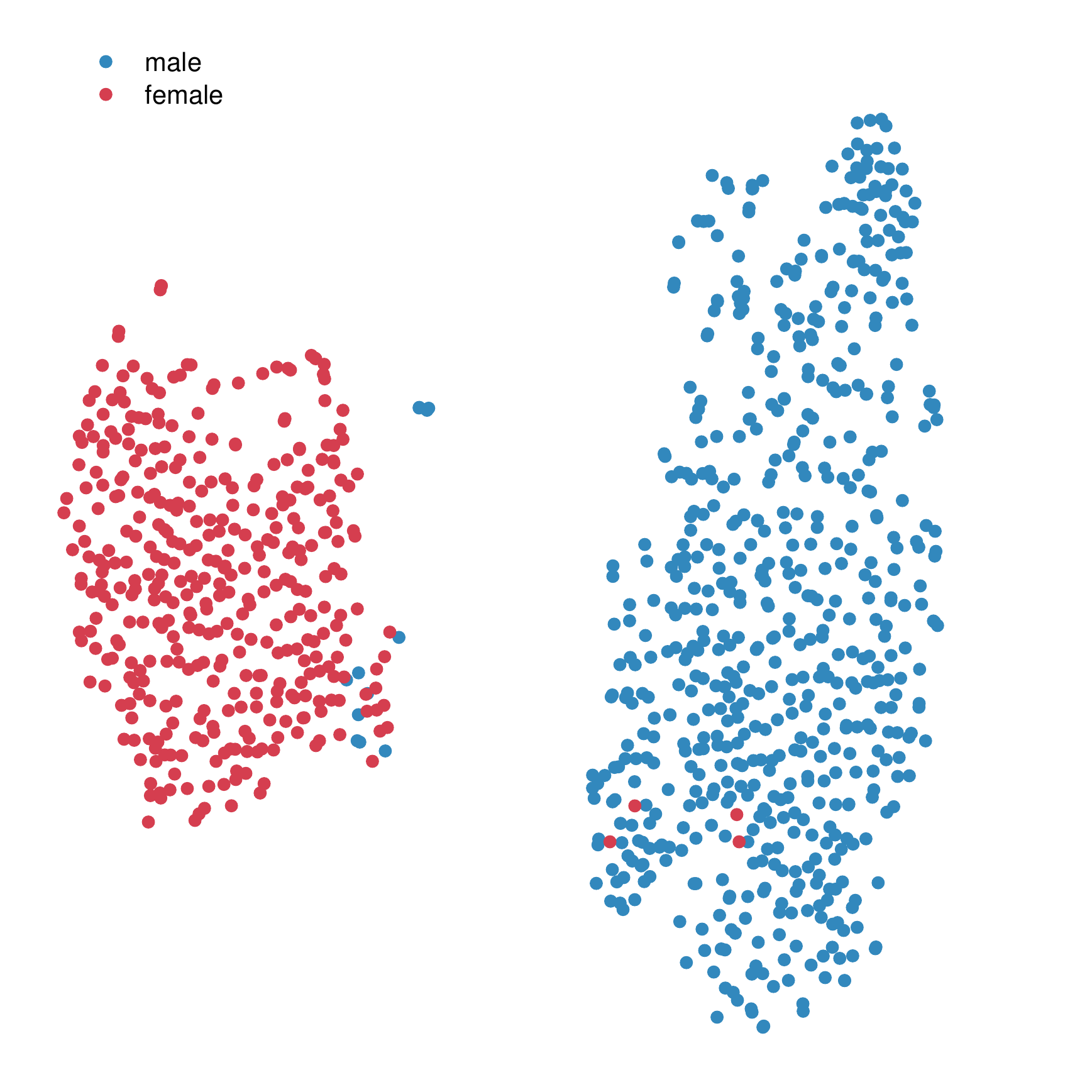}\quad%
  \includegraphics[width=0.25\linewidth]{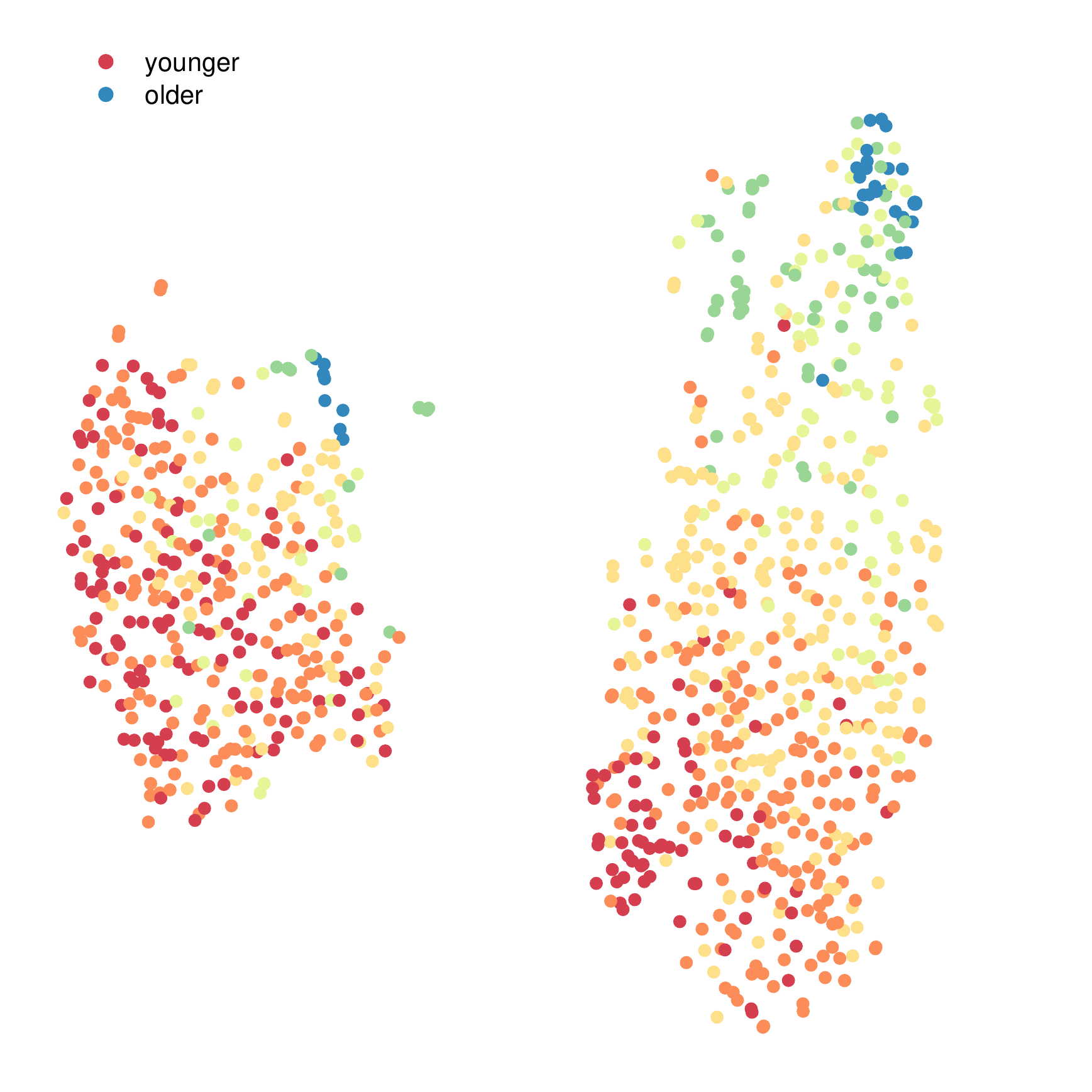}\\
  \vspace{-.2cm}
  \parbox               {0.25\linewidth}{\centering\scriptsize{}(a) Face; identity}\quad%
  \parbox               {0.25\linewidth}{\centering\scriptsize{}(b) Face; gender}\quad%
  \parbox               {0.25\linewidth}{\centering\scriptsize{}(c) Face; age}\\
  \includegraphics[width=0.25\linewidth]{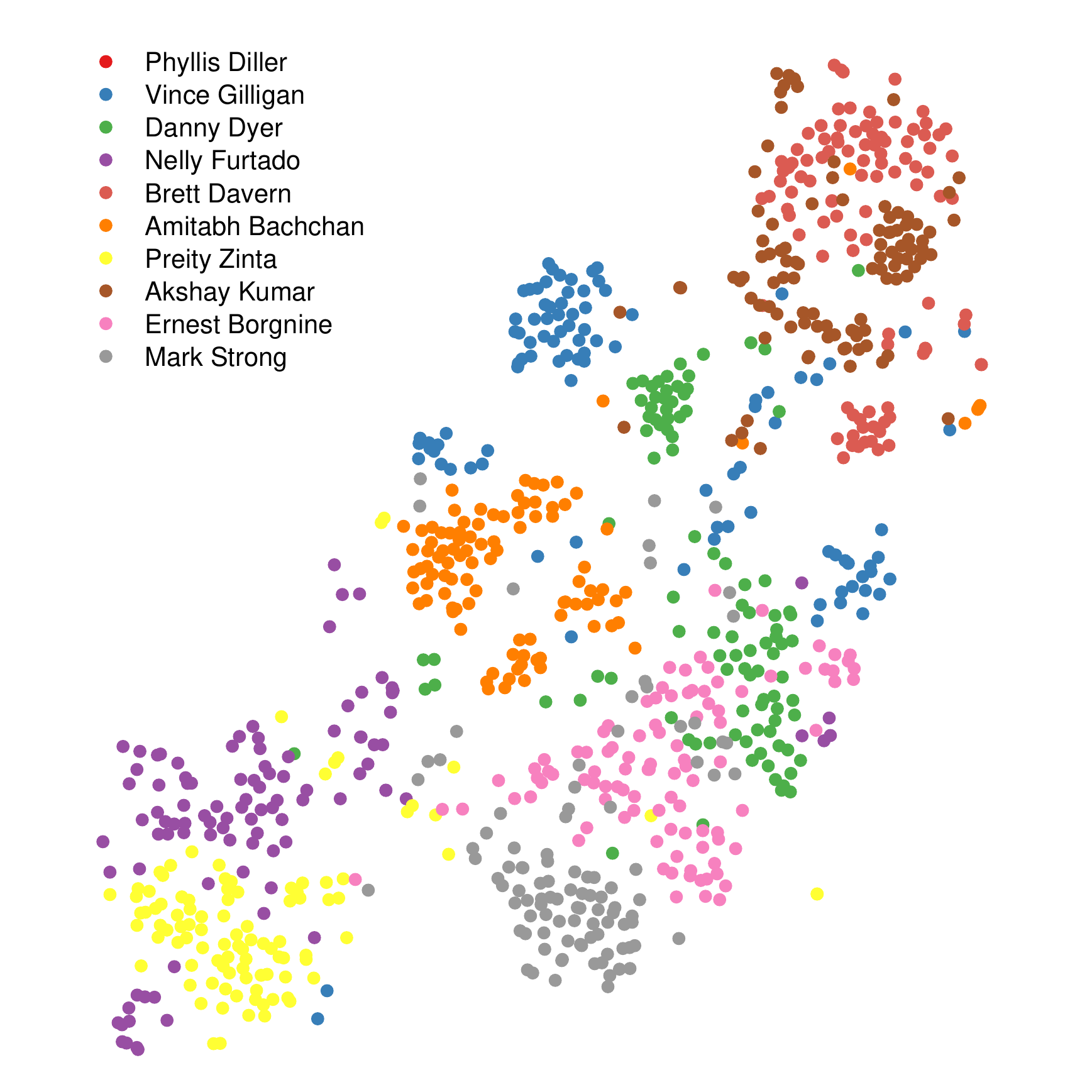}\quad%
  \includegraphics[width=0.25\linewidth]{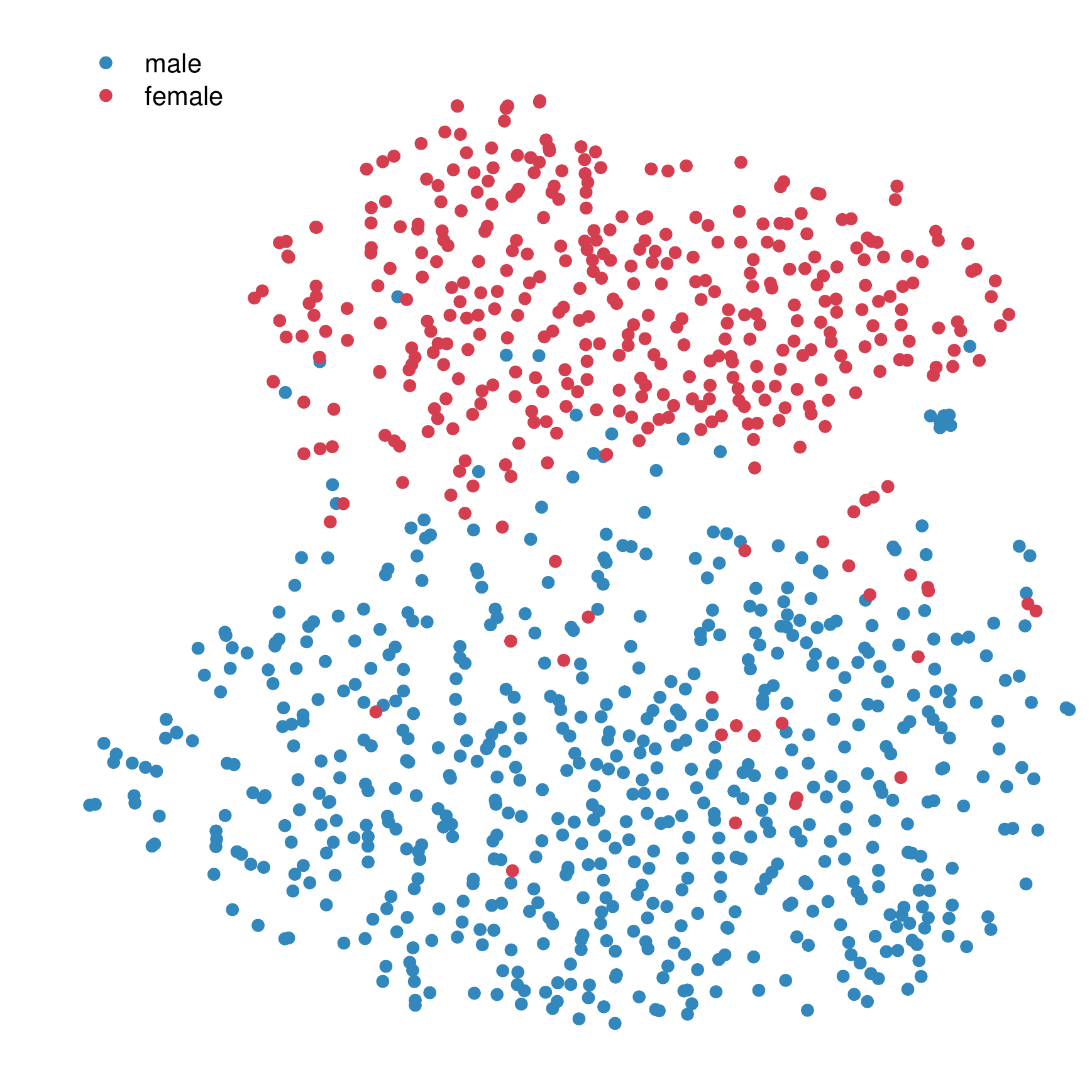}\quad%
  \includegraphics[width=0.25\linewidth]{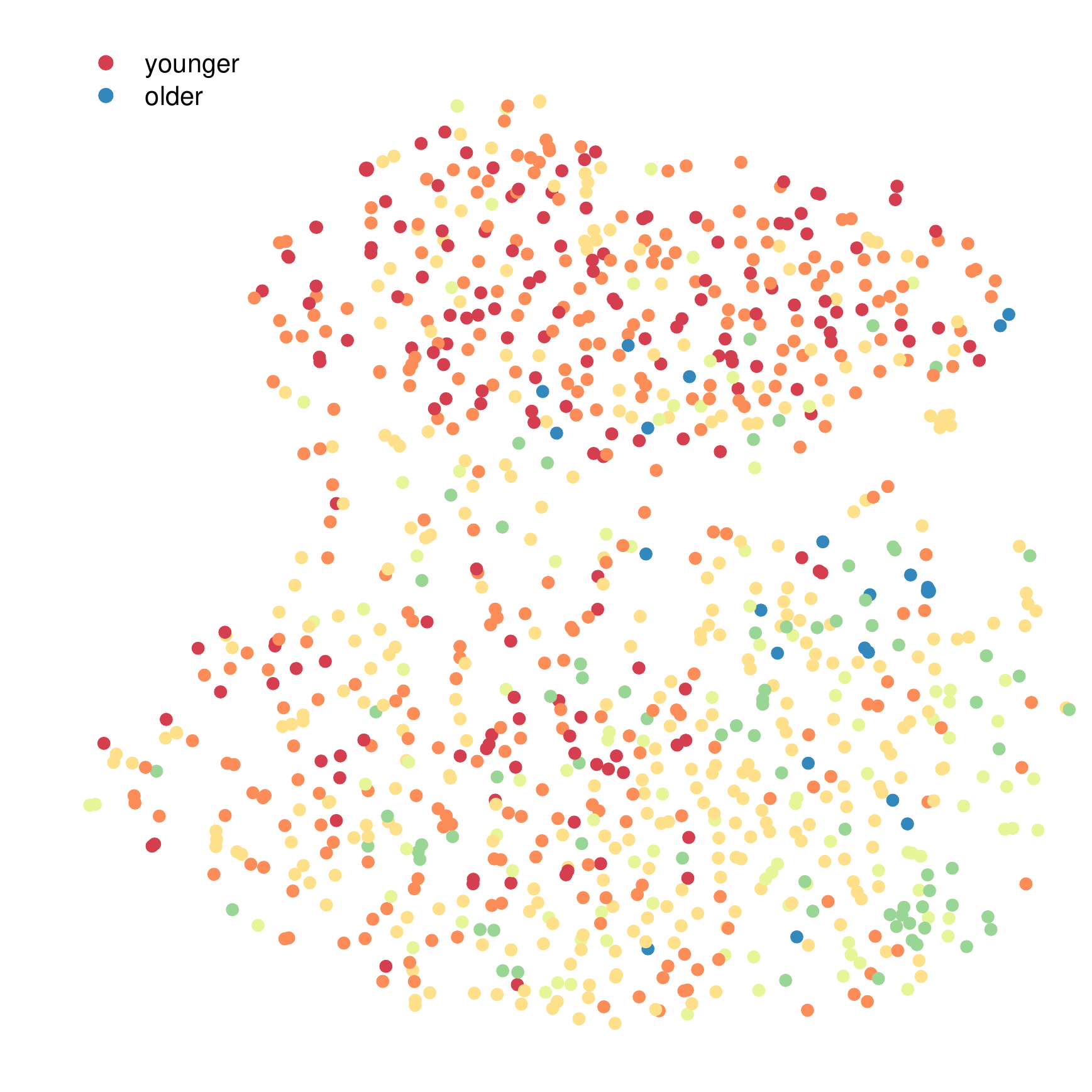}\\
  \vspace{-.2cm}
  \parbox               {0.25\linewidth}{\centering\scriptsize{}(d) Voice; identity}\quad%
  \parbox               {0.25\linewidth}{\centering\scriptsize{}(e) Voice; gender}\quad%
  \parbox               {0.25\linewidth}{\centering\scriptsize{}(f) Voice; age}\\
  \vspace{-.1cm}
  \caption{The t-SNE visualization of face (a--c) and voice (d--f) representations, color-coded with the identity and demographic information. Age (c,f) color-codes continuous values, the legend showing only the two extremes; the rest categorical values. No demographic attribute was used to train representations.}
  \label{fig:tsne}
  \vspace{-.4cm}
\end{figure}
}

\newcommand{\addAttributel}{%
\begin{table}[t]
  \caption{Analysis of the information encoded in face and voice representations. We report the mean average precisions (mAP) with 99$\%$ confidence intervals (CI) obtained from 20 trials of holdout cross validations. Those having a CI overlapping the random chance with a 5$\%$ margin ($50\pm5\%$) are marked in red. Performance not higher than random suggests that the representation is not distinctive enough for that classification task. Again, none of these attributes were used for training.}
  \vspace{-.2cm}
  \resizebox{\linewidth}{!}{%
    \setlength{\tabcolsep}{1.5mm}%
    \begin{tabular}{@{}lcccccccccccccc@{}}%
      \toprule
      \multicolumn{2}{@{}l}{\multirow{2}{*}[-.5mm]{Modality}} & \multirow{2}{*}[-.5mm]{Gender} & \multirow{2}{*}[-.5mm]{Fluency} & \multicolumn{5}{c}{Age} & \multicolumn{6}{c}{Ethnicity} \\
      \cmidrule(lr){5-9} \cmidrule(ll){10-15}
      && & &  $<$30 &  30s &  40s &  50s &  $\geq$60 &  1 &  2 &  3 &  4 &  5 &  6\\
      \midrule
      \multirow{2}{*}{Face repr.} & mAP & 99.4 & 65.4 & 76.8 & \hlck{60.7} & \hlck{59.1} & 71.9 & 81.9 & 84.5 & 82.5 & 84.6 & 74 & 72 & 67.3\\
      &  CI & ${\pm}$0.2 & ${\pm}$7.9 & ${\pm}$4.5 & \hlck{${\pm}$8.9} & \hlck{${\pm}$7.6} & ${\pm}$4.2 & ${\pm}$7.0 & ${\pm}$11.6 & ${\pm}$5.2 & ${\pm}$5.3 & ${\pm}$5.6 & ${\pm}$8.1 & ${\pm}$11.1\\
      \cmidrule{1-15}
      \multirow{2}{*}{Voice repr.} & mAP & 90.4 & \hlck{53.9} & \hlck{60.6} & \hlck{53.3} & \hlck{50.8} & \hlck{53} & \hlck{59.8} & 84.7 & 69.6 & \hlck{53.3} & \hlck{58.2} & \hlck{53.8} & 63.8\\ 
      &  CI & ${\pm}$4.0 & \hlck{${\pm}$3.8} & \hlck{${\pm}$7.5} & \hlck{${\pm}$3.4} & \hlck{${\pm}$0.6} & \hlck{${\pm}$4.1} & \hlck{${\pm}$5.7} & ${\pm}$9.3 & ${\pm}$7.5 & \hlck{${\pm}$3.4} & \hlck{${\pm}$5.1} & \hlck{$\pm$4.7} & $\pm$7.1\\
      \bottomrule
    \end{tabular}%
  }%
  \label{tab:attributel}
  \vspace{-.4cm}
\end{table}
}

\newcommand{\addProsodic}{%
\begin{figure}[t]
  \centering
  \includegraphics[width=0.24\linewidth]{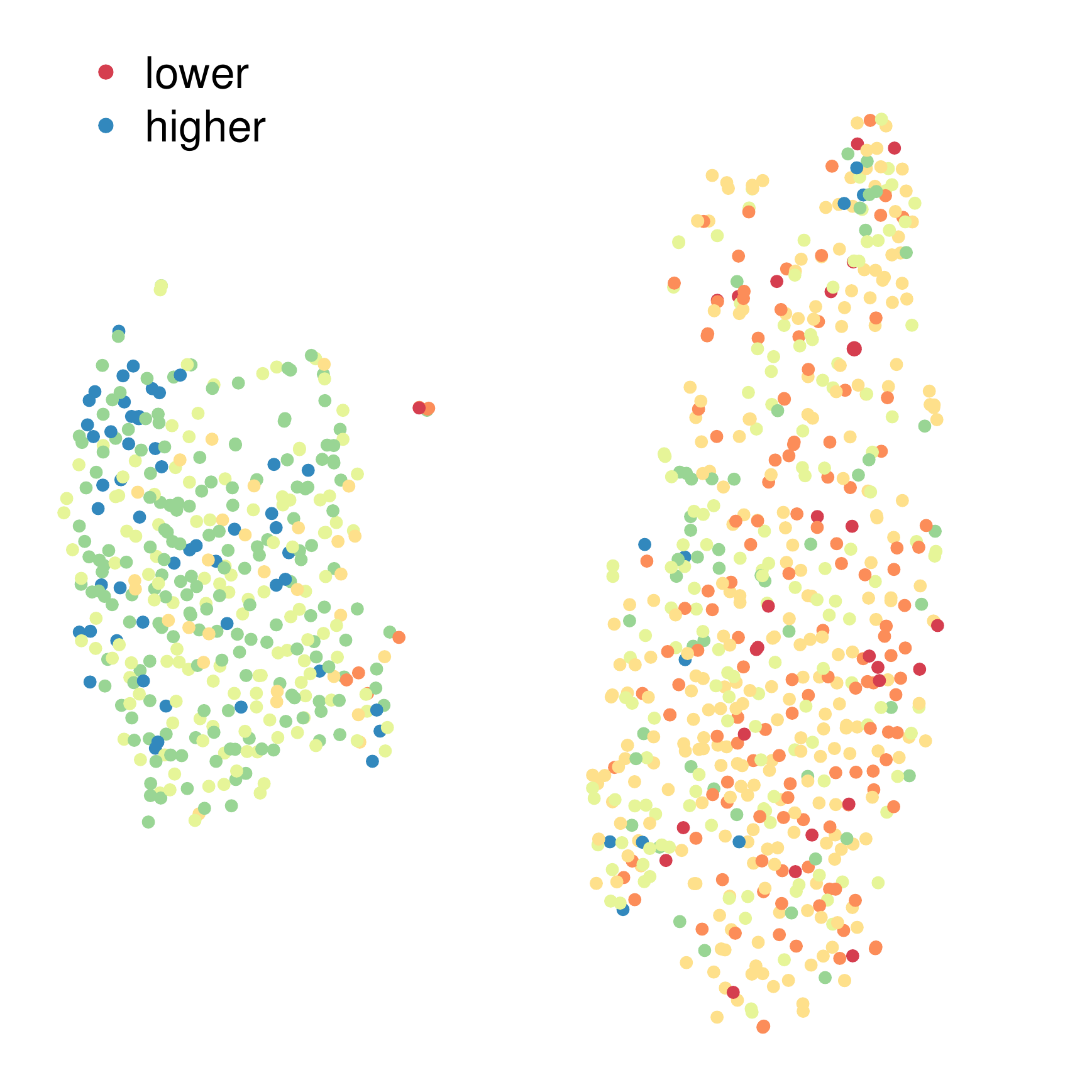}\hfill
  \includegraphics[width=0.24\linewidth]{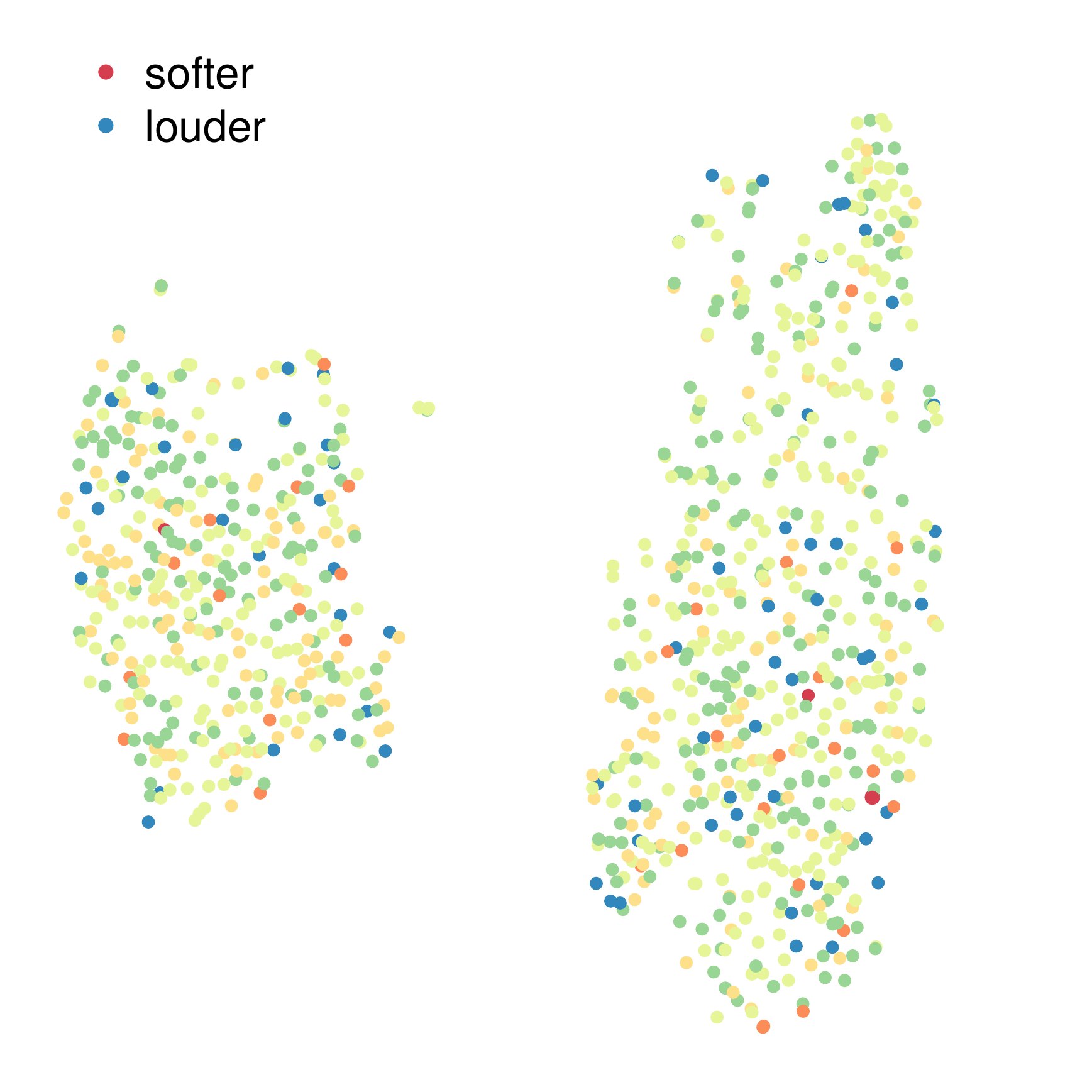}\hfill
  \includegraphics[width=0.24\linewidth]{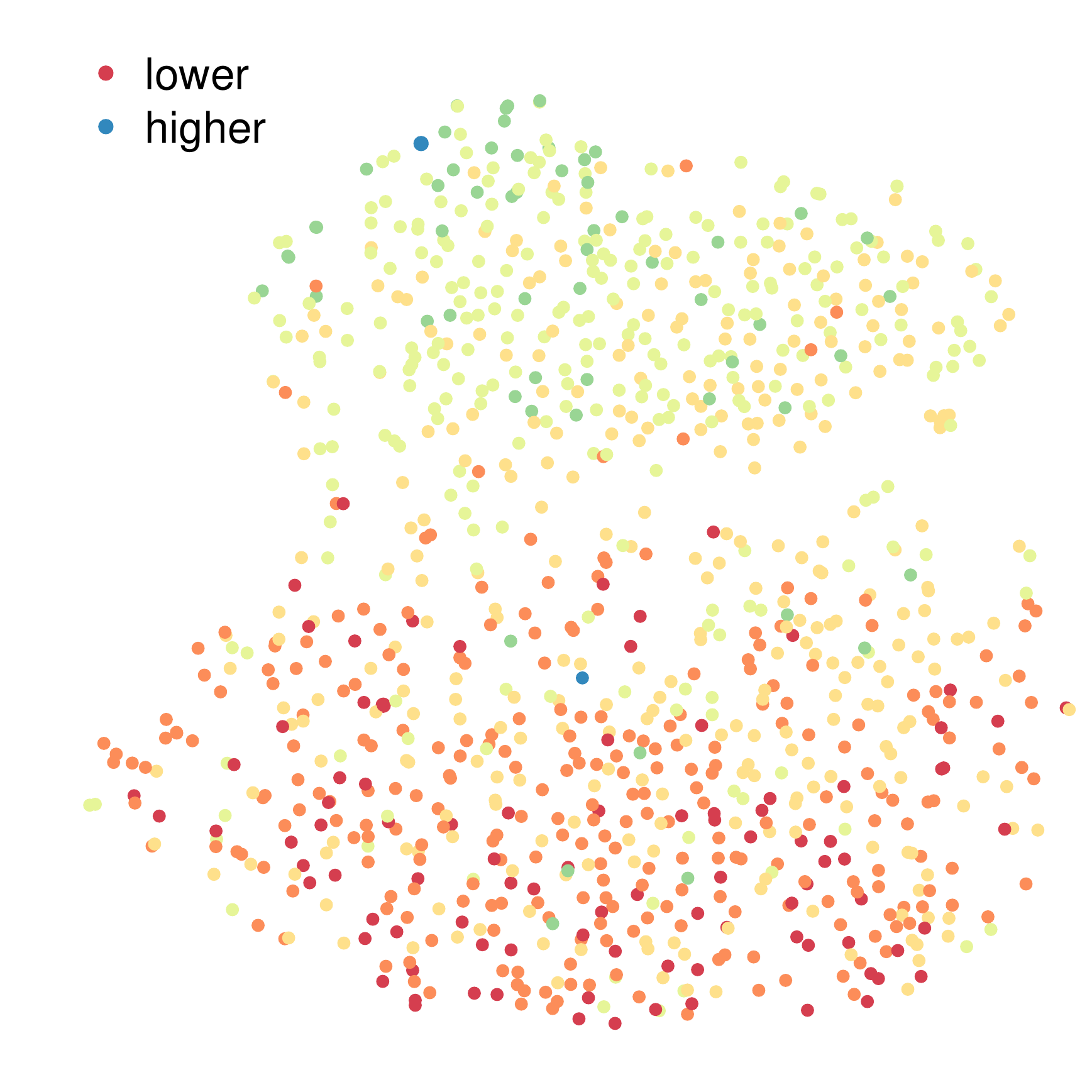}\hfill
  \includegraphics[width=0.24\linewidth]{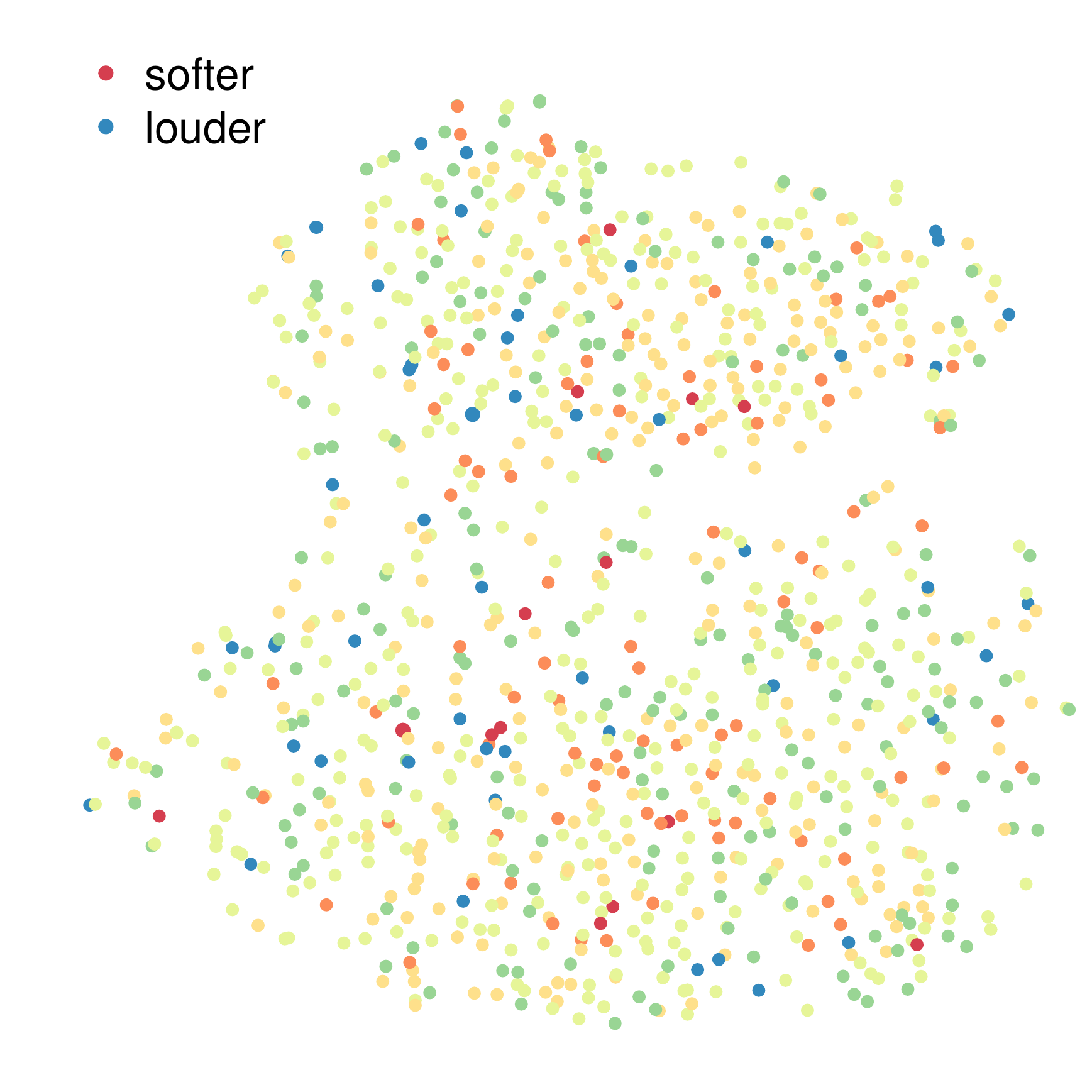}\\
  \vspace{-.2cm}
  \parbox               {0.24\linewidth}{\centering\scriptsize{}(a) Face; pitch}\hfill
  \parbox               {0.24\linewidth}{\centering\scriptsize{}(b) Face; loudness}\hfill
  \parbox               {0.24\linewidth}{\centering\scriptsize{}(c) Voice; pitch}\hfill
  \parbox               {0.24\linewidth}{\centering\scriptsize{}(d) Voice; loudness}
  \vspace{-.3cm}
  \caption{The t-SNE visualization of face (a,b) and voice (c,d) representations with respect to two prosodic features, voice pitch and loudness.}
  \label{fig:prosodic}
  \vspace{-.4cm}
\end{figure}
}

\newcommand{\addPitchGEFAEmbed}{%
  \begin{wrapfigure}{r}{0.23\linewidth}
  \centering
  \hspace{-1mm}%
  \includegraphics[width=\linewidth]{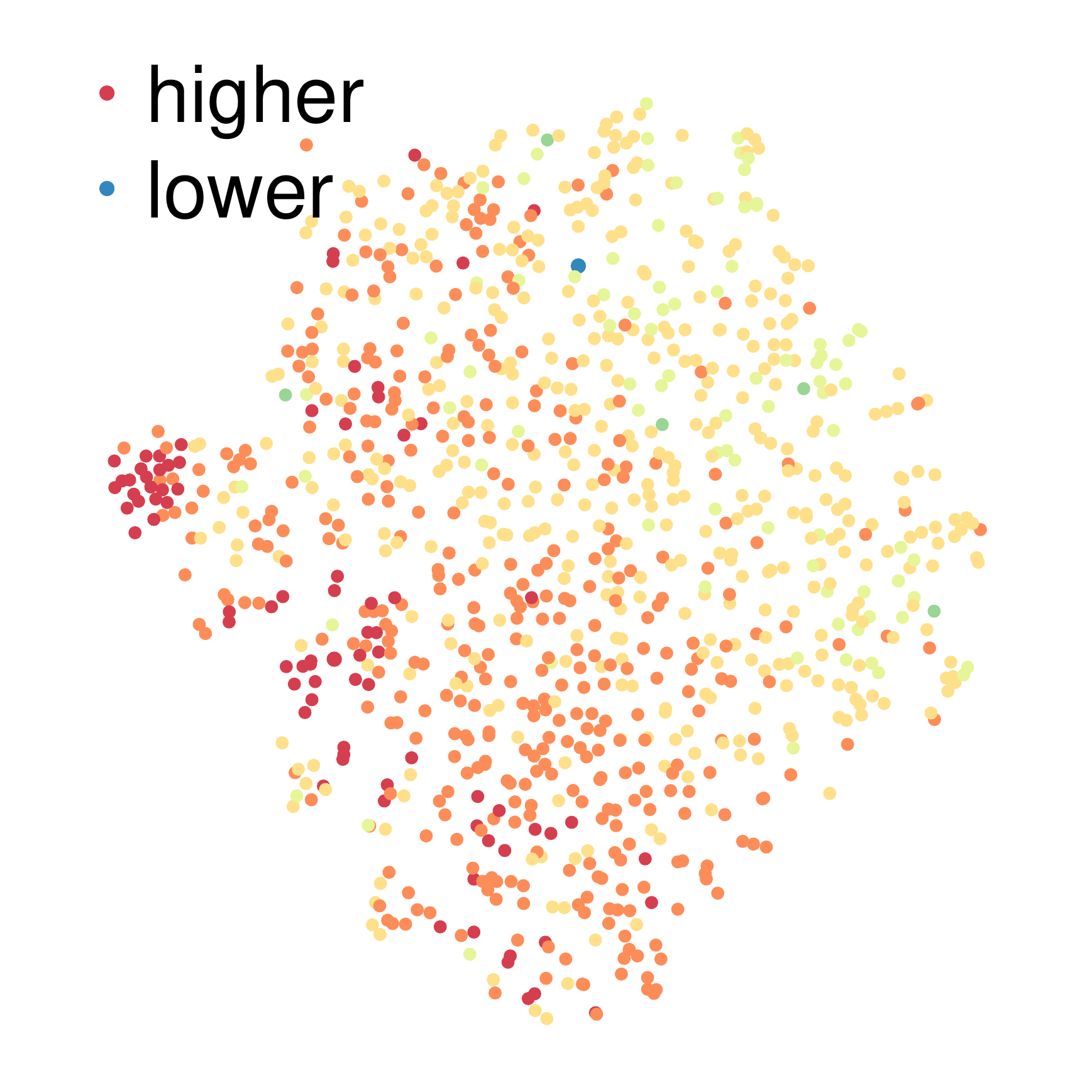}\\
  \vspace{-3mm}
  \hspace{-1mm}%
  \parbox{\linewidth}{\centering\scriptsize{}Voice pitch}
  \end{wrapfigure}%
}

\newcommand{\addFacialEmbed}{%
\begin{wraptable}{r}{0.69\linewidth}%
  \vspace{-.4cm}
  \caption{Facial features encoded in our representation. Classification precisions for select CelebA features are shown.}
  \vspace{.1cm}
  \resizebox{\linewidth}{!}{%
    \centering%
    \small%
    \setlength{\tabcolsep}{1.5mm}%
    \begin{tabular}{@{}lllll@{}}
      \toprule
      Modality & Big nose & Chubby & Double chin & Baldness \\
      \midrule
      Face repr. & 62.9 ${\pm}$\,7.7 & 69.4 ${\pm}$\,4.7 & 69.4 ${\pm}$\,5.2 & 81.0 ${\pm}$\,5.8 \\
      Voice repr. & 50.4 ${\pm}$\,7.5 & 57.2 ${\pm}$\,4.3 & 61.5 ${\pm}$\,9.7 & 71.0 ${\pm}$\,10.2 \\ 
      \bottomrule
    \end{tabular}%
  }%
  \label{tab:facial}
  \vspace{-.1cm}
\end{wraptable}
}

\newcommand{\addRecall}{%
\begin{table}[b]
  \vspace{-.3cm}
  \centering
  \caption{Results of cross-modal retrieval on the VoxCeleb test set. 25,000 samples in the VoxCeleb test set are divided into sets of the following sizes and the recall was averaged. Each R@$K$ denotes recall@$K$. Random indicates the recall of random guess.}
  \vspace{-.2cm}
  \scriptsize
  \begin{tabular}{@{}%
      L{0.10\linewidth}%
      R{0.10\linewidth}%
      R{0.10\linewidth}%
      R{0.10\linewidth}%
      R{0.10\linewidth}%
      R{0.10\linewidth}%
      R{0.10\linewidth}%
      @{}}
    \toprule
    Direction                        & Set size & R@1 & R@5 & R@10 & R@50 & R@100 \\
    \midrule
    \multirow{3}{*}{V $\rightarrow$ F} & 250      &   3.5\%      &  11.3\%      &  15.1\%      &  51.3\%      &  84.4\%  \\
                                     & 1,000    &   1.9\%      &   7.2\%      &  13.7\%      &  41.3\%      &  61.8\%  \\
                                     & 5,000    &   2.3\%      &   7.7\%      &  12.7\%      &  34.8\%      &  47.0\%  \\
    \cmidrule{1-7}
    \multirow{3}{*}{F $\rightarrow$ V} & 250      &   2.7\%      &   8.6\%      &  12.3\%      &  52.6\%      &  82.6\%  \\
                                     & 1,000    &   3.3\%      &  10.7\%      &  18.1\%      &  45.2\%      &  65.7\%  \\
                                     & 5,000    &   0.7\%      &   7.2\%      &  13.9\%      &  42.6\%      &  61.2\%  \\
    \cmidrule{1-7}
    \multicolumn{2}{c}{Random}                  &   0.4\%      &   2.0\%      &   4.0\%      &  20.0\%      &  40.0\%  \\
    \bottomrule
  \end{tabular}
  \label{tab:recall}
\end{table}
}

\newcommand{\addComparisons}{%
\begin{table}[t]
  \centering
  \caption{Performance with alternative model components (for V $\rightarrow$ F). The results consistently support the learnability of the associations with comparable performances.}
  \vspace{-.2cm}
  \scriptsize
  \begin{tabular}{@{}%
      L{0.45\linewidth}%
      L{0.10\linewidth}%
      L{0.10\linewidth}%
      L{0.10\linewidth}%
      L{0.10\linewidth}%
      L{0.10\linewidth}%
      @{}}
    \toprule
    \multirow{2}{*}[-.5mm]{Model configurations} & \multicolumn{5}{c}{Demographic grouping} \\
    \cmidrule{2-6}
                                   & -- & G & E & G/E & G/E/F/A \\
    \midrule
    Siamese net with same subnets  & 76.5\% & 59.9\% & 76.2\% & 60.7\% & 57.2\% \\
    Triplet net with VGG-Vox \& VGG-Face & 81.9\% & 67.3\% & 81.8\% & 66.7\% & 57.5\% \\
    Classification network         & 77.6\% & 62.2\% & 77.4\% & 61.6\% & 58.4\% \\
    Our model                      & 78.2\% & 62.9\% & 76.4\% & 61.6\% & 59.0\% \\
    \bottomrule
  \end{tabular}
  \label{tab:comparisons}
  \vspace{-.5cm}
\end{table}
}

\newcommand{\addGeneralizationTableAndFigureSeparatelyUsingMinipage}{%
\begin{figure}[t]
  \begin{minipage}{0.59\linewidth}
    \vspace{-2mm}
    \captionof{table}{Performance of our model measured on the dataset collected for our user studies (\secref{userstudy}).}
    \label{tab:generalization}
    \vspace{2mm}
    \resizebox{\linewidth}{!}{%
      \scriptsize
      \begin{tabular}{@{}%
          L{0.17\linewidth}%
          C{0.13\linewidth}%
          C{0.13\linewidth}%
          C{0.13\linewidth}%
          C{0.13\linewidth}%
          C{0.175\linewidth}%
          @{}}
        \toprule
        \multirow{2}{*}[-.5mm]{Direction} & \multicolumn{5}{c}{Demographic grouping} \\
        \cmidrule(ll){2-6}
                                    & -- & G & E & G/E & G/E/F/A \\
        \midrule
        V $\rightarrow$ F           & 71.2\% & 58.0\% & 70.7\% & 55.0\% & 42.6\% \\
        F $\rightarrow$ V           & 64.5\% & 52.4\% & 65.1\% & 52.4\% & 51.4\% \\
        \bottomrule
      \end{tabular}%
    }%
  \end{minipage}%
  \hfill%
  \begin{minipage}{0.37\linewidth}
    \raisebox{-.5\height}{\includegraphics[width=0.49\linewidth]{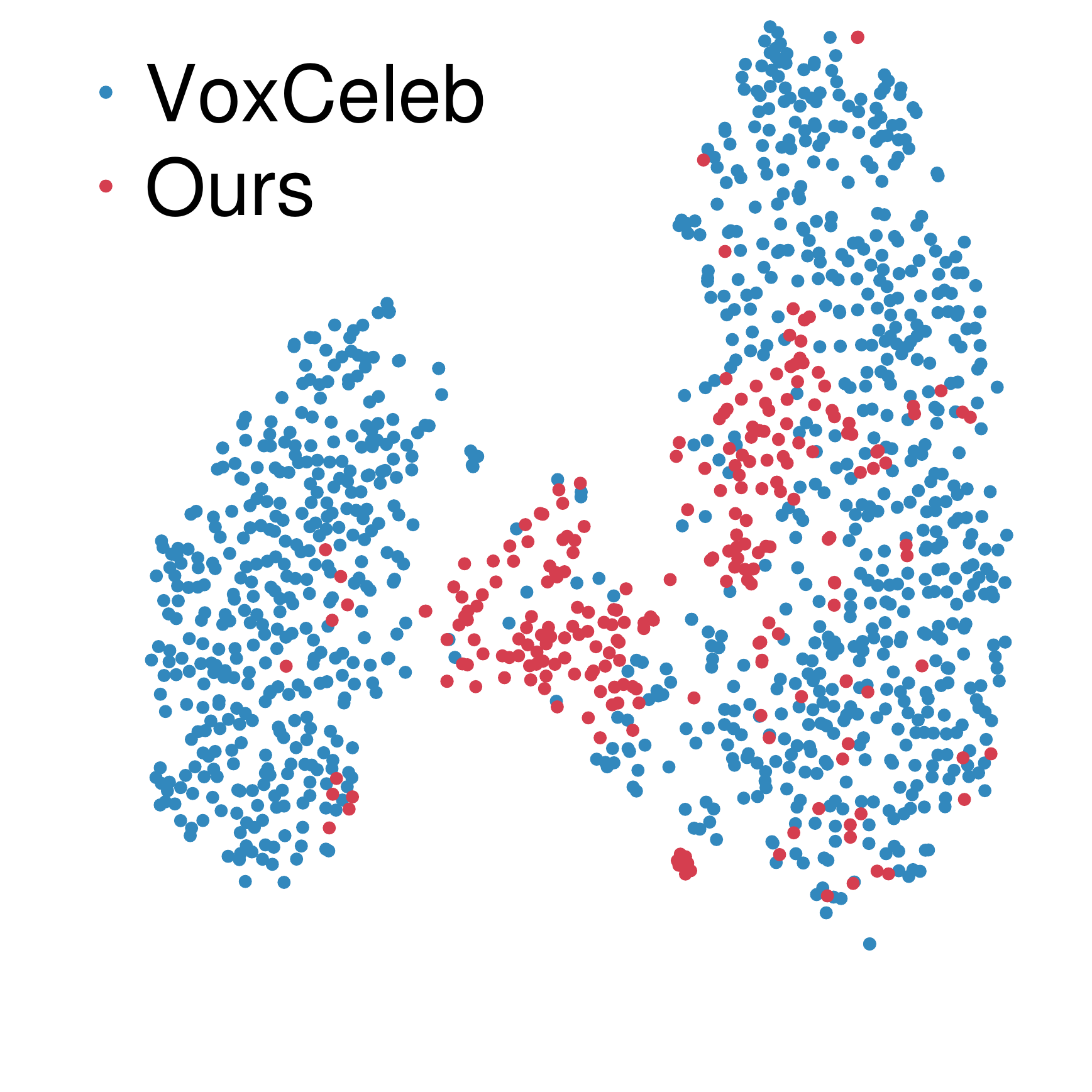}}\hfill%
    \raisebox{-.5\height}{\includegraphics[width=0.49\linewidth]{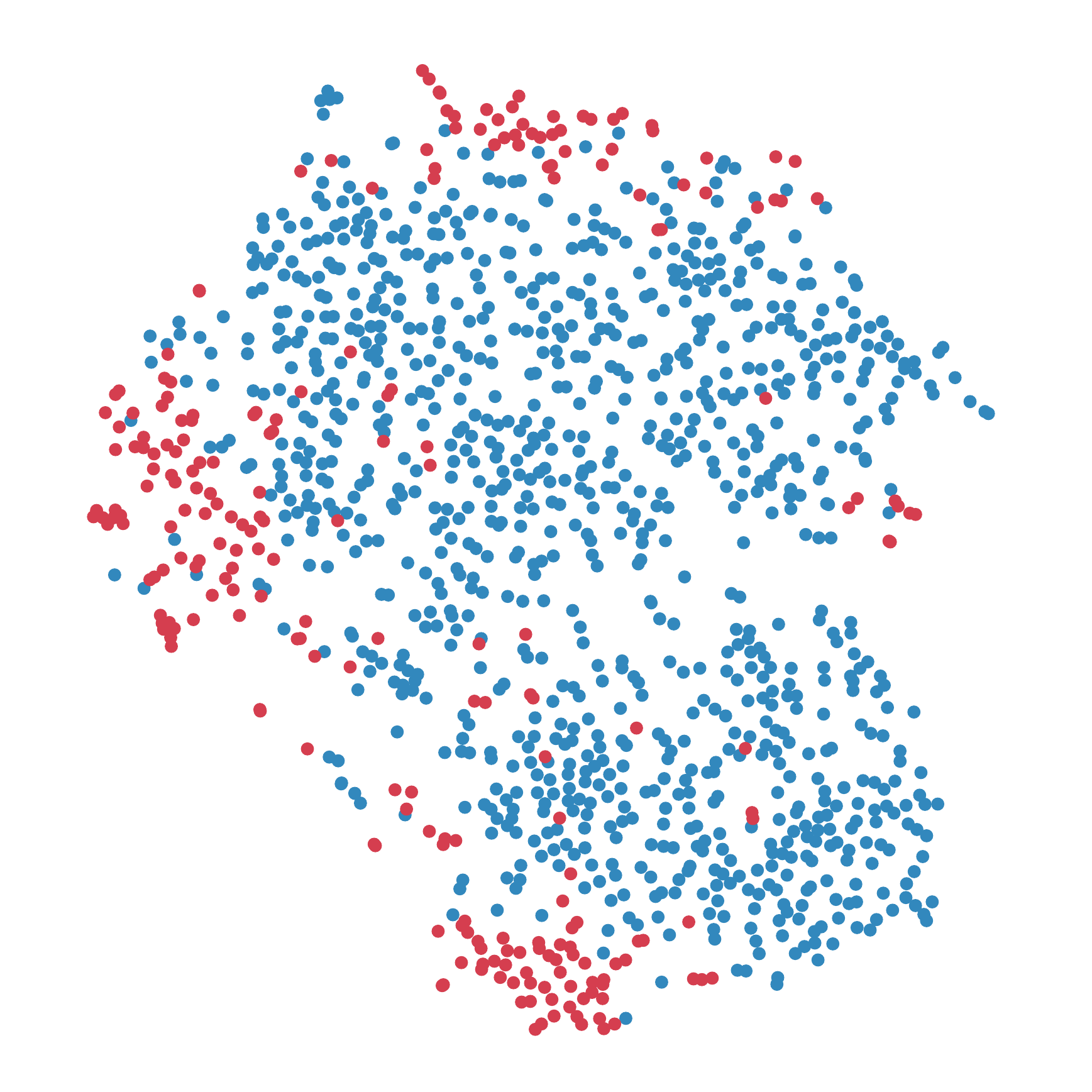}}\\
    \vspace{-4mm}
    \parbox{0.49\linewidth}{\centering\scriptsize{}(a) Face dist.}\hfill%
    \parbox{0.49\linewidth}{\centering\scriptsize{}(b) Voice dist.}\\
    \vspace{-2mm}
    \caption{Distributions of VoxCeleb and our dataset samples.}
    \label{fig:sampledistributions}
  \end{minipage}
  \vspace{-.4cm}
\end{figure}
}

%% file: intro.tex
\section{Introduction}
\label{sec:intro}

\indent\indent\emph{``Can machines put a face to the voice?''}\\[-.15cm]

\noindent
We humans often deduce various, albeit perhaps crude, information from the voice of others, such as gender, approximate age and even personality.
We even imagine the appearance of the person on the other end of the line when we phone a stranger.
Can machines learn such human ability?
In this paper we pose questions about whether machines can put faces to voices, or vice versa, like humans presumably do, and if they can, how accurately they can do so.
To answer these questions, we need to define what ``putting a face to a voice'' means. We approximate this task by designing a simple discrete test:
we judge whether a machine can choose the most plausible facial depiction of the voice it hears, given multiple candidates. This definition has a number of advantages: (1)~it is easy to implement in machines, (2)~it is possible to conduct the same test on human subjects, and (3)~the performance can be quantitatively measured.

Neuroscientists have observed that the multimodal associations of faces and voices play a role in perceptual tasks such as speaker recognition~\cite{Kriegstein2005,Joassin2011,Zweig2015}.
Recently, the problem was brought to the computer vision community and it has been shown that such ability can be implemented by machine vision and intelligence~\cite{Nagrani2018}.
We perform experimental studies both on human subjects and machine models. Compared to prior human-subject studies, we collect a new, larger dataset consisting of audiovisual recordings of human speeches performed by non-celebrity individuals with more diverse demographic distributions, on which human-subject study is conducted to set a more accurate baseline for human performances.
Unlike the prior computational model~\cite{Nagrani2018}, which models the task as an $n$-way classification, we learn the overlapping information between the two modalities, inspired by the findings of neuroscientists. This allows us to analyze both modalities in the same embedding space by measuring the distance between two modal representations directly, which enables cross-modal retrieval. We analyze what information we have learned, and examine potential connections between our learned representation and modal features of faces and voices alone. We show that our representation has a close connection to certain demographic attributes such as age and gender, some facial features, and prosodic features like voice pitch. We expect our approach to further lead to new opportunities for cross-modal synthesis and editing.

\vspace{-.2cm}

\paragraph{Contributions.}

Our technical contributions include the following.
\begin{itemize}
	\item We provide an extensive human-subject study, with both the participant pool and dataset larger and more diverse than those used in prior studies, where we verify that humans are capable of correctly matching unfamiliar face images to corresponding voice recordings and vice versa with greater than chance accuracy. We provide a statistical analysis with diverse controls on demographic attributes and various levels of homogeneity of studied groups.
	\item We learn the co-embedding of modal representations of human faces and voices, and evaluate the learned representations extensively, revealing unsupervised correlations to demographic, prosodic, and facial features. We compare a number of existing techniques to learn the representation and show that we obtain consistent performances, independent of particular computational models, on the matching task on a par with human performance.
	\item We present a new dataset of the audiovisual recordings of speeches by 181 individuals with diverse demographic background, totaling over 3 hours of recordings, with the demographic annotations.
\end{itemize}

\vspace{-.2cm}

\paragraph{Limitations.}

While we use our own dataset for human-subject studies, we use an existing dataset of celebrities (the VoxCeleb dataset~\cite{Nagrani2017}) to train our computational model, due to the two experiments' respective characteristics and practical concern about data collection. 
Humans have prior knowledge about celebrities, which can affect their performance on VoxCeleb, while the deep neural network we use requires a large amount of data, rendering our dataset short of scale.
Further, conducting user studies on such a huge dataset would also require a comparably large number of test participants. 
Thus, it should be avoided to compare the numbers directly between the two studies; rather, the results should be understood such that both humans and our computational model achieve statistically significant, better than random performances.
Collecting a large dataset of non-celebrity audiovisual recordings comparable to VoxCeleb in size is an important and non-trivial task which we leave to future work.

%% file: related.tex
\section{Related Work}
\label{sec:related}

Studies on face-voice association span multiple disciplines. Among the most relevant to our work are cognitive science and neuroscience, which study human subjects, and machine learning, specifically, cross-modal modeling.

\vspace{-.2cm}

\paragraph{Human capability for face-voice association.}

Behavioural and neuroimaging studies of face-voice integration show clear evidence of early perceptual integrative mechanisms between face and voice processing pathways. The study of Campanella and Belin~\cite{campanella2007integrating} reveals that humans leverage the interface between facial and vocal information for both person recognition and identity processing.
This human capability is unconsciously learned by processing a tremendous number of auditory-visual examples throughout their whole life~\cite{gaver1993world}, and the ability to learn the associations between faces and voices starts to develop as early as three-months old~\cite{brookes2001three}, without intended discipline.\footnote{In machine learning terminology, this could be seen as natural supervision~\cite{Owens2016visually} or self-supervision~\cite{doersch2015unsupervised} with unlabeled data.}
This ability has also been observed in other primates~\cite{sliwa2011spontaneous}.

These findings led to the question about to what extent people are able to correctly match which unfamiliar voice and face belong to the same person~\cite{Kamachi2003,Lachs2004,Mavica2013,smith2016matching}.
Early work~\cite{Kamachi2003,Lachs2004} argued that people could match voices to dynamically articulating faces but not to static photographs.
More recent findings of Mavica and Barenholtz~\cite{Mavica2013} and Smith et al.~\cite{smith2016matching} contradicted these results, and presented evidence that humans can actually match \emph{static} facial images to corresponding voice recordings with greater than chance accuracy. In a separate study, Smith et al.\ also showed that there is a strong agreement between the participants' ratings of a model's femininity, masculinity, age, health and weight made separately from faces and voices~\cite{smith2016concordant}. 
The discrepancy between these sets of studies were attributed to the different experimental procedures. For instance, Kamachi et al.~\cite{Kamachi2003} and Lachs and Pisoni~\cite{Lachs2004} presented the stimuli sequentially (participants either heard a voice and then saw two faces or saw a face and then heard two voices), while the latter works presented faces and voices simultaneously. In addition, the particular stimuli used could also have led to a difference in performance. For example, Kamachi et al.\ experimented with Japanese models, whereas Marvica and Barenholtz used Caucasian models. Smith et al.~\cite{smith2016matching} showed that different models vary in the extent to which they look and sound similar, and performance could be highly dependent on the particular stimuli used.

The closest work to our human subject study is Mavica and Barenholtz's experiment. We extend the previous work in several ways. First, we exploit crowdsourcing to collect a larger and more diverse dataset of models. We collected faces and voices of 181 models of different gender, ethnicity, age-group and first-language. This diversity allowed us to investigate a wider spectrum of task difficulties according to varying control factors in demographic parameters. Specifically, whereas previous work only tests on models from a homogenous demographic group (same gender, ethnicity, age group), we vary the homogeneity of the sample group in each experiment and test models from same gender~(G), same gender and ethnicity~(G/E), same gender, ethnicity, first language and age group~(G/E/F/A). By comparing the performances across experiments, we explicitly test the assumption, hereto taken for granted, that people infer demographic information from both face and voice and use this to perform  the matching task. 

\vspace{-.2cm}

\paragraph{Audiovisual cross-modal learning by machinery.}

Inspired by the early findings from cognitive science and neuroscience that humans integrate audiovisual information for perception tasks~\cite{mcgurk1976hearing,jones1975eye,Shelton1980}, the machine learning community has also shown increased interest in the visual-auditory cross-modal learning.
The key motivation has been to understand whether machine learning models can reveal correlations across different modalities.
With the recent advance of deep learning, multi-modal learning leverages neural networks to mine common or complementary information effectively from large-scale paired data.
In the real world, the concurrency of visual and auditory information provides a natural supervision~\cite{Owens2016ambient}.
Recent emergence of deep learning has witnessed the understanding of the correlation between audio and visual signals in applications such as:
improving sound classification~\cite{Arandjelovic2017} by combining images and their concurrent sound signals in videos; scene and place recognition~\cite{Aytar2016} by transferring knowledge from visual to auditory information;
vision-sound cross modal retrieval~\cite{Owens2016visually,Owens2016ambient,Soler2016}; and sound source localization in visual scenes~\cite{Senocak2018}.
These works focus on the fact that visual events are often positively correlated with their concurrent sound signals. This fact is utilized to learn representations that are modality-invariant.
We build on these advances and extend to the face-voice pair.

Nagrani et al.~\cite{Nagrani2018} recently presented a computational model for the face-voice matching task. While they see it as a binary decision problem, we focus more on the shared information between the two modalities and extract it as a representation vector residing in the shared latent space, in which the task is modeled as a nearest neighbor search.
Other closely related work include Ngiam et al.~\cite{Ngiam2011} and Chung et al.~\cite{Chung2017}, which showed that the joint signals from face and audio help disambiguate voiced and unvoiced consonants. Similarly, Hoover et al.~\cite{hoover2017putting} and Gebru et al.~\cite{gebru2015tracking} developed systems to identify active speakers from a video by jointly observing the audio and visual signals. Although the voice-speaker matching task seems similar, these work mainly focus on distinguishing active speakers from non-speakers at a given time, and they do not try to learn cross-modal representations. 
A different line of work has also shown that recorded or synthesized speech can be used to generate facial animations of animated characters~\cite{Taylor2017,Karras2017audio} or real persons~\cite{Suwajanakorn2017}. 

Our interest is to investigate whether people look and sound similar, i.e., to explore the existence of the learnable relationship between the face and voice.
To this end, we leverage the face-voice matching task. We examine whether faces and voices encode redundant identity information and measure to which extent.

%% file: human.tex
\section{Study on Human Performance}
\label{sec:userstudy}

We conducted a series of experiments to test whether people can match a voice of an unfamiliar person to a static facial image of the same person. Participants were presented with photographs of two different models and a 10-second voice recording of one of the models. They were asked to choose one and only one of the two faces they thought would have a similar voice to the recorded voice (V $\rightarrow$ F). We hypothesized that people may rely on information such as gender, ethnicity and age inferred from both face and voice to perform the task. To test this possibility, in each experiment, we added additional constraints on the sample demography and only compared models of the same gender (G - Experiment 1), same gender and ethnicity (G/E - Experiment 2), and finally same gender, ethnicity, first language, and age group (G/E/F/A - Experiment 3), specifically male pairs and female pairs from non-Hispanic white, native speakers in their 20s. For the most constrained condition (G/E/F/A), we also performed a follow-up experiment, where participants were presented with a single facial image and two voice recordings and chose the recording they thought would be similar to the voice of the person in the image (F $\rightarrow$ V).

\subsection{Dataset}

While there exist multiple large-scale audiovisual datasets of human speakers, notably in the context of speech or speaker recognition~\cite{Nagrani2017,Chung2017}, they contain widely known identities, such as celebrities or public figures.
Thus, for our human subject study, we used Amazon Mechanical Turk to collect a separate dataset consisting of 239 video clips of 181 unique non-celebrities. Participants recorded themselves through their webcam, while reading out short English scripts.
In addition to the video recordings, participants fill out a survey about their demographic information: gender, ethnicity, age, and their first language.
The demographic distribution of the acquired dataset is tabulated in \tabref{demography}.
See our supplementary material for acquisition details and the accompanying dataset to examine samples.

\subsection{Protocol}

For the face-voice matching experiments, we conducted a separate study also through Amazon Mechanical Turk. Before starting the experiment, participants filled out a questionnaire about their demographic information, identical to the one above presented for data collection. Following the questionnaire, they completed 16 matching tasks, along with 4 control tasks for quality control. Each task consists of comparing two pairs of faces and selecting one of them as matching a voice recording (vice versa for Experiment 4). Two of the four control tasks check for consistency; we repeat a same pair of faces and voice. The other two control for correctness; we add two pairs with one male model and one female model. From preliminary studies we noticed that people are generally very good at identifying gender from face or voice, and indeed less than 3\% of the participants incorrectly answered the correctness control questions (11 out of 301 participants). In the analysis, we discarded data from participants who failed in two or more control questions (9/301).

The rest of the 16 tasks comprise of 16 different pairs. Each unique person in the dataset is paired with 8 other persons from the dataset, randomly selected within the experiment's demographic constraint (Experiment 1: same gender, Experiment 2: same gender and ethnicity, Experiments 3 and 4: same gender, ethnicity, age group and first language). Each participant in the experiment was presented with 16 randomly selected pairs (8 male pairs and 8 female pairs). The pairs were presented sequentially. Participants had to listen to the audio recording(s) and choose an answer, before they could move on to the next pair. No feedback was given on whether their choice was correct or not, precluding learning of face-voice pairings. We also discarded data from participants who partook in the data collection (4/301).

\subsection{Results}

\addHumanPerformance

\tabref{human_performance} shows the average performance across participants for each of the four experimental conditions. Individual $t$ tests found significantly better than chance performance (50\%) for each of the four experimental conditions. An ANOVA comparing the four experiments found a significant difference in performance ($F = 21.36$, $p < 0.001$). Tukey's HSD showed that performance in Experiment 1~(G) was significantly better than Experiment 2~(G/E) ($p < 0.05$), and performance in Experiment 2~(G/E) was significantly better than Experiment 3~(G/E/F/A) ($p < 0.05$). However, results from Experiment 3~(V $\rightarrow$ F) and Experiment 4~(F $\rightarrow$ V) were not significantly different from one another.

\addModelPerformanceEmbed
Similarly to Mavica and Barenholtz~\cite{Mavica2013}, in order to assess whether some models were more or less difficult to match, for Experiment 3, we also calculated the percentage of trials on which the participants chose the correct response whenever the model's face was presented either as the correct match or as the incorrect comparison. In other words, a high score for a model means participants are able to correctly match the model's face to its voice as well as reject matching that face to another person's voice.
Shown above is the average performance for each of the 42 models (18 male and 24 female) in Experiment 3, sorted by performance. Despite the wide variance in performance, we observe a clear trend toward better-than-chance performance, with 34 of the 42 models (80\%) yielding a performance above 50\%.

Overall, participants were able to match a voice of an unfamiliar person to a static facial image of the same person at better than chance levels. The performance drop across experimental conditions 1 to 3 supports the view that participants leverage demographic information inferred from the face and voice to perform the matching task. Hence, participants performed worse when comparing pairs of models from demographically more homogeneous groups. This was an assumption taken for granted in previous work, but not experimentally tested. More interestingly, even for the most constrained condition, where participants compared models of the same gender, ethnicity, age group and first language, their performance was better than chance. This result aligns with that of Mavica and Barenholtz~\cite{Mavica2013} that humans can indeed perform the matching task with greater than chance accuracy even with static facial images. The direction of inference (F $\rightarrow$ V vs.\ V $\rightarrow$ F) did not affect the performance.

%% file: machine.tex
\section{Cross-modal Metric Learning on Faces and Voices}
\label{sec:network}

\newcommand{\voicenet}{f_\mathrm{V}}
\newcommand{\facenet}{f_\mathrm{F}}
\newcommand{\voice}{\mathbf{v}}
\newcommand{\face}{\mathbf{f}}
\newcommand{\negface}{\face^-}
\newcommand{\posface}{\face^+}
\newcommand{\dist}{d}
\newcommand{\negdist}{\dist^-}
\newcommand{\posdist}{\dist^+}
\newcommand{\softmax}{\text{softmax}}
\newcommand{\loss}{\mathcal{L}}

Our attempt to learn cross-modal representations between faces and voices is inspired by the significance of the overlapping information in certain cognitive tasks like identity recognition, as discussed earlier. We use standard network architectures to learn the latent spaces that represent the visual and auditory modalities for human faces and voices, respectively, and are compatible enough to grasp the associations between them.
Analogous to human unconscious learning~\cite{gaver1993world}, we train the networks to learn the voice-face pairs from naturally paired face and voice data without other human supervision.

\subsection{Network Architecture}
\label{sec:arch}

The overall architecture is based on the triplet network~\cite{Hoffer2015}, which is widely used for metric learning.
As subnetworks for two modalities, we use VGG16~\cite{Simonyan2014} and SoundNet~\cite{Aytar2016}, which have shown sufficient model capacities while allowing for stable training in a variety of applications.
In particular, SoundNet was devised in the context of transfer learning between visual and auditory signals.

Unlike typical triplet configurations where all three subnetworks share the weights, in our model, two heterogeneous subnetworks are hooked up to the triplet loss. 
The face subnetwork $\facenet$ is based on VGG16, where the \texttt{conv5\_3} layer is average-pooled globally, resulting in 512-d output. It is fed to a 128-d fully connected layer with the ReLU activation, followed by another 128-d fully connected layer but without ReLU, which yields the face representation. The voice subnetwork $\voicenet$ is based on SoundNet, whose \texttt{conv6} layer is similarly average-pooled globally, yielding 512-d output. It is then fed to two fully-connected layers with the same dimensions as those in the face subnetwork one after another.
In our experiments with the voice as the reference modality, for a single voice subnetwork, there are two face subnetworks with shared weights.

During training, for each random voice sample $\voice$, one positive face sample $\posface$ and one negative sample $\negface$ are drawn, and the tuple $(\voice, \posface, \negface)$ is fed forward to the triplet network. Optimizing for the triplet loss
\begin{equation}
  \loss(\voice, \posface, \negface) = \left\| \, \softmax ([\posdist, \negdist]) - [0, 1] \, \right\|_2^2
\end{equation}
minimizes the $L_2$ distance between the representations of the voice and the positive face, $\posdist = \| \voicenet(\voice) - \facenet(\posface) \|_2$, while maximizing the $L_2$ distance between those of the voice and the negative face, $\negdist = \| \voicenet(\voice) - \facenet(\negface) \|_2$, pushing representations of the same identity closer and pulling those of different identities away.

\subsection{Dataset}
\label{sec:voxceleb_dataset}

\addDemography

Our collected dataset of 239 samples was not large enough to train a large deep neural network. Thus, we turned to unconstrained, ``in-the-wild'' datasets, which provide a large amount of videos mined from video sharing services like YouTube.
We use the VoxCeleb dataset~\cite{Nagrani2017} to train our network. From the available 21,063 videos, 
114,109 video clips of 1,251 celebrities are cut and used. We split these into two sets: randomly chosen 1,001 identities as the training set and the rest 250 identities as the test set. The dataset comes with facial bounding boxes. We first filtered the bounding boxes temporally as there were fluctuations in their sizes and positions, and enlarged them by 1.5 times to ensure that the entire face is always visible. From each clip, the first frame and first 10 seconds of the audio are used, as the beginning of the clips is usually well aligned with the beginning of utterances.
We manually annotated the samples in the test set with demographic attributes, which allowed us to conduct the experiments with the same controls as presented in \secref{userstudy} and to examine the clustering on such attributes naturally arising in the learned representations (\secref{visualization}).
The demographic distributions of the annotated test set are illustrated in \tabref{demography}.

\subsection{Training}

All face images are scaled to 224$\times$224 pixels. Audio clips are resampled at 22,050~Hz and trimmed to 10 seconds; those shorter than 10 seconds are tiled back to back before trimming.
Training tuples are randomly drawn from the pool of faces and voices: for a random voice, a random but distinct face of the same identity and a random face of a different identity are sampled.
We use Adam~\cite{Kingma2014} to optimize our network with $\beta_1 = 0.9$ and $\beta_2 = 0.999$, and the batch size of 8. We use the pretrained models of VGG16 and SoundNet. The fully connected layers are trained from scratch with a learning rate of $10^{-3}$ and the pretrained part of the network is fine-tuned with a learning rate of $10^{-5}$ at the same time. The training continues for 240k iterations, while the learning rates are decayed by a factor of $10^{-1}$ after every 80k iterations. After 120k iterations, the network is trained with \emph{harder} training samples, where 16 tuples are sampled for each batch, from which only the 8 samples with the highest losses are used for back-propagation.
See the supplementary material for more details.
We train a separate model for each direction: a V $\rightarrow$ F network with the voice as the reference modality 
and an F $\rightarrow$ V network with the face as the reference. 

\subsection{Results}

\addMachinePerformance

We conducted the same experiments introduced in \secref{userstudy} using our trained model. A voice recording is fed to the network along with two candidate face images, resulting in three representation vectors.
Then the face candidate closer to the voice in the representation space in $L_2$ metric is picked as the matching face.
The performance of our computational model is tabulated in \tabref{machine_performance}. 

Similarly to our user study, we measure the test accuracy on a number of different conditions. We replicate the conditions of Experiments~1~(G), 2~(G/E), and 3~(G/E/F/A) as before but in both directions (thus including Experiment~4), in addition to two more experiments where the accuracy is measured on the same ethnic group~(E) and on the entire test set samples~(--).
For Experiment~3~(G/E/F/A), we show the accuracy on the single, largest homogeneous group of people in the test set (non-Hispanic white, male native speakers in their 30s). Note that we used the age group of 30s instead of 20s, which were the largest group in our user study dataset, as the VoxCeleb test set demography includes more identities in their 30s. These largest groups are marked in boldface in \tabref{demography}.

We observe that the gender of the subject provides the strongest cue for our model to decide the matching, as we assume it does for human testers.\footnote{Gender is such a strong cue that we use it for control questions in our user study. See \secref{userstudy}.} Unlike the experiments with human participants, conditioning on ethnicity lowers the accuracy only marginally. For the most constrained condition (G/E/F/A) the accuracy shows about 20\% drop from the uncontrolled experiment.

These results largely conform to our findings from the user study (\tabref{human_performance}).
One noticeable difference is that the performance drop due to the demographic conditioning is less drastic in the machine experiments~($\sim$4\%) than in the human experiments~($\sim$13\%), while their accuracies on the most controlled group (G/E/F/A; i.e., the hardest experiment) are similar (59.0\% and 58.4\%, respectively). Note that the accuracy on the uncontrolled group was not measured on human participants, and the machine's best accuracy should not erroneously be compared to the human best accuracy, which is already measured among same gender candidates.

\subsection{Evaluations of the Learned Representation}
\label{sec:visualization}

\addTSNE

\figref{tsne} demonstrates the clustering that emerges in our learned representation using the t-SNE visualization~\cite{VanDerMaaten2008}.
The samples in the t-SNE plots are colored so as to denote particular attribute values associated with them (from either their identities or our annotations) and visualize the attribute distribution in the feature space. Bear in mind that both the t-SNE and our network have not seen any such demographic attributes during training, and that at no point has the association between the attributes and our learned representations been introduced to the network. This allows us an unbiased assessment of the attributes' correlation to the feature distribution.
We drew 100 random samples for each of 10 unique identities from the VoxCeleb test set for the identity visualization in \figref{tsne}ad, which shows per-identity clustering.
Additionally, we drew 1,000 random samples for demographic attribute visualizations in \figref{tsne}bcef.
The learned representation forms the clearest clusters regarding gender (\figref{tsne}be), which explains the performance drop when the experiment is constrained by gender. Also noticeable is the distribution by age (\figref{tsne}cf). While correlated with gender, it shows a distinct grouping to gender, in particular for face representations.
The t-SNE visualization does not reveal similar clustering with respect to the first language or the ethnic group (shown in the supplementary material).

\addAttributel

In \tabref{attributel}, we further evaluate our representation using linear classifiers trained on our representations.
We examine whether or not any additionally interpretable information is encoded in the representations, and how much discrepancy there exists between the representations from two modalities.
Following the data-driven probing used in Bau et al.~\cite{bau2017network}, we use the demographic attributes as probing data to see how accurately they can be predicted from our representations. 
Given the set of representations and their corresponding attributes, we train one-vs-all SVM classifiers for each attribute.
The results further support that, while the attributes are never used for training, the learned representation encodes a significant amount of attribute information.
They also demonstrate that our representation encodes additional information, more prominent in the face modality. Statistical insignificance of the age group classification from voice representations aligns with the t-SNE (\figref{tsne}f), which shows less obvious patterns than those found in its face counterpart (\figref{tsne}c).
See our supplementary material for more visualizations and further evaluations.

\addPitchGEFAEmbed

We show t-SNE with two prosodic features to examine a potential correlation between vocal features and our learned representations. 
In \figref{prosodic}, our representation forms clusters with respect to voice pitch (fundamental frequency), while it does not with respect to voice loudness. 
Shown on the right is the t-SNE of G/E/F/A--controlled samples, color-coded with their voice pitch: the voice pitch is found in our learned representations even in the most controlled sample group. This shows that our representation remains informative about voice pitch, which presumably is one of the \emph{residual signals} beyond demographic attributes and affects the performance, among many possible factors.
We trained SVM on our representation to predict CelebA attributes~\cite{Liu2015} and show in \tabref{facial} several attributes suggesting correlation.{\parfillskip0pt\par}

\addProsodic

\addFacialEmbed

\noindent This demonstrates that our representation encodes certain information related to these attributes without supervision. Each cell shows the mean average precision with 99\% confidence intervals. The correlation with baldness reveals the attribute's strong gender bias.
Like demographic attributes, neither prosodic nor facial features were used for training.

Lastly, we use the learned representations for cross-modal retrieval. Given a face (voice) sample, we retrieve the voice (face) samples closest in the latent space. We report recall@$K$ in \tabref{recall}, which measures the portion of queries where the true identity is among the top $K$ retrievals, as in Vendrov et al.~\cite{Vendrov2015}, for varying $K$ and set sizes. The number of samples per identity was kept the same while samples within the same identity was randomly chosen.

\addRecall

\subsection{Discussions}

\paragraph{Comparisons and model parameter selection.}

We experimented the task with a number of different model components: a Siamese network with the same subnetworks, trained with a contrastive loss~\cite{Chopra2005}; the same triplet network but with VGG-Vox~\cite{Nagrani2017} as voice subnetwork and VGG-Face~\cite{Parkhi2015} as face subnetwork; and finally a binary classification network inspired by the ``$L^3$ network''~\cite{Arandjelovic2017}, an audiovisual correlation classifier, and Nagrani et al.'s model~\cite{Nagrani2018}, which determines, given a face and a voice, whether or not the two belong to the same identity.
For the classification network, the positive class probability is used as a measurement of the similarity between a face and a voice to determine the matching candidate.
The networks are trained in a similar manner to the network presented in \secref{arch} with hyper-parameters manually tuned to ensure the best possible performances.
The results shown in \tabref{comparisons} present similar performance on our experiments, which supports the \emph{learnability} of the overlapping information between two modalities regardless of the particular network architecture.
We also measured the test accuracy with varying configurations of the presented network, e.g., the dimensions of the fully connected layers (and thus those of the resulting representation vectors). While this did not influence the test accuracy much, generally smaller (narrower) fully connected layers resulted in a better performance.
We detail the comparisons with different architectures and choices of hyper-parameters in more details in our supplementary material.

\addComparisons

\vspace{-.2cm}

\paragraph{Representation asymmetry.} 

We observed that face and voice representations are learned asymmetrically, depending on the modality used as the reference of the triplet.
We simply trained two networks, one with the voice as reference for voice-to-face retrieval, and vice versa. A more sophisticated model, such as the quadruplet network~\cite{Chen2017}, could be used to alleviate this issue. In this work, however, we focus more on showing the feasibility of the task using widely-used models, minimizing the complexity and dependency on a particular architecture.

\vspace{-.2cm}

\paragraph{Cross-domain generalization.}

\tabref{generalization} indicates that our model trained on the VoxCeleb dataset results in lower performance on our dataset used for user studies---a phenomenon known as ``dataset bias''~\cite{Torralba2011}.
The t-SNE's of the samples drawn from both datasets in \figref{sampledistributions} show that the distributions of VoxCeleb and our dataset do not exactly overlap, which is more prominent for voices: while the faces in our dataset seem to be covered by those of VoxCeleb, the voices tend to be outside of the gamut of the VoxCeleb sample distribution. 
This is likely attributed to the fact that VoxCeleb is collected from published interviews with professional-quality audio whereas our dataset consists of webcam recordings.
It is also suggested that the appearance of celebrities is more diverse and gender-typical than non-celebrities, and its distribution is not dense enough in the regions where most non-celebrity faces are distributed.
This could be alleviated by additional fine-tuning on the new dataset or by domain adaptation (e.g., Tzeng et al.~\cite{Tzeng2017}), which is left to future work.

\addGeneralizationTableAndFigureSeparatelyUsingMinipage

\vspace{-.2cm}

\paragraph{Accent and regional cues.}

It is worth noting that cultural or regional cues, such as accent or the subject's appearance, can play a role in the face-voice matching task. In fact, both face and voice contain rich information about a person's identity that cannot be controlled or separated out in a simple way (e.g., it is difficult to imagine a ``completely neutral face'' devoid of racial, emotional or personality cues). Instead of attempting to factor out all such cues, we use self-reported demographic information to control for the most common and objective identity factors. We manually filtered data with very strong accent or background noise.
In the most homogeneous group (controlled by all G/E/F/A), we compare native speakers to minimize the influence of accents. We also cropped images to the facial bounding boxes to minimize subtle hints from background.

%% file: conclusion.tex
\section{Conclusion}
\label{sec:conclusion}

We studied the associations between human faces and voices through a series of experiments: first, with human subjects, showing the baseline for how well people perform such tasks, and then on machines using deep neural networks, demonstrating that machines perform on a par with humans.
We expect that our study on these associations can provide insights into challenging tasks in a broader range of fields, pose fundamental research challenges, and lead to exciting applications.

For example, cross-modal representations such as ours could be used for finding the voice actor that \emph{sounds} like how an animated character \emph{looks}, manipulating synthesized facial animations~\cite{Taylor2017,Karras2017audio,Suwajanakorn2017} to harmonize with corresponding voices, or as an entertaining application to find the celebrity whose voice sounds like a user's face or vice versa.
While understanding of the face-voice association at this stage is far from perfect, its advance could potentially lead to additional means towards criminal investigation like lie detection~\cite{Wu2017}, which is still arguable but practically used. However, we emphasize that, similar to lie detectors, such associations should not be used for screening purposes or as hard evidence.
Our work suggests the possibility of learning the associations by referring to a part of the human cognitive process, but not their definitive nature, which we believe would be far more complicated than it is modeled as in this work.

%% file: acknowledgments.tex
\paragraph{Acknowledgments.}
This work was funded in part by the QCRI--CSAIL computer science research program. Changil Kim was supported by a Swiss National Science Foundation fellowship P2EZP2 168785. We thank Sung-Ho Bae for his help.

%% file: supp_content.tex
\section{Data Collection for Human Performance}

Both the data acquisition and user study were carried out through web applications deployed via Amazon Mechanical Turk. In the following, we present further details of the two tasks.

\subsection{User Study}

\begin{figure}
  \centering
  \includegraphics[width=\linewidth,trim={2cm 1.5cm 2cm 1cm},clip]{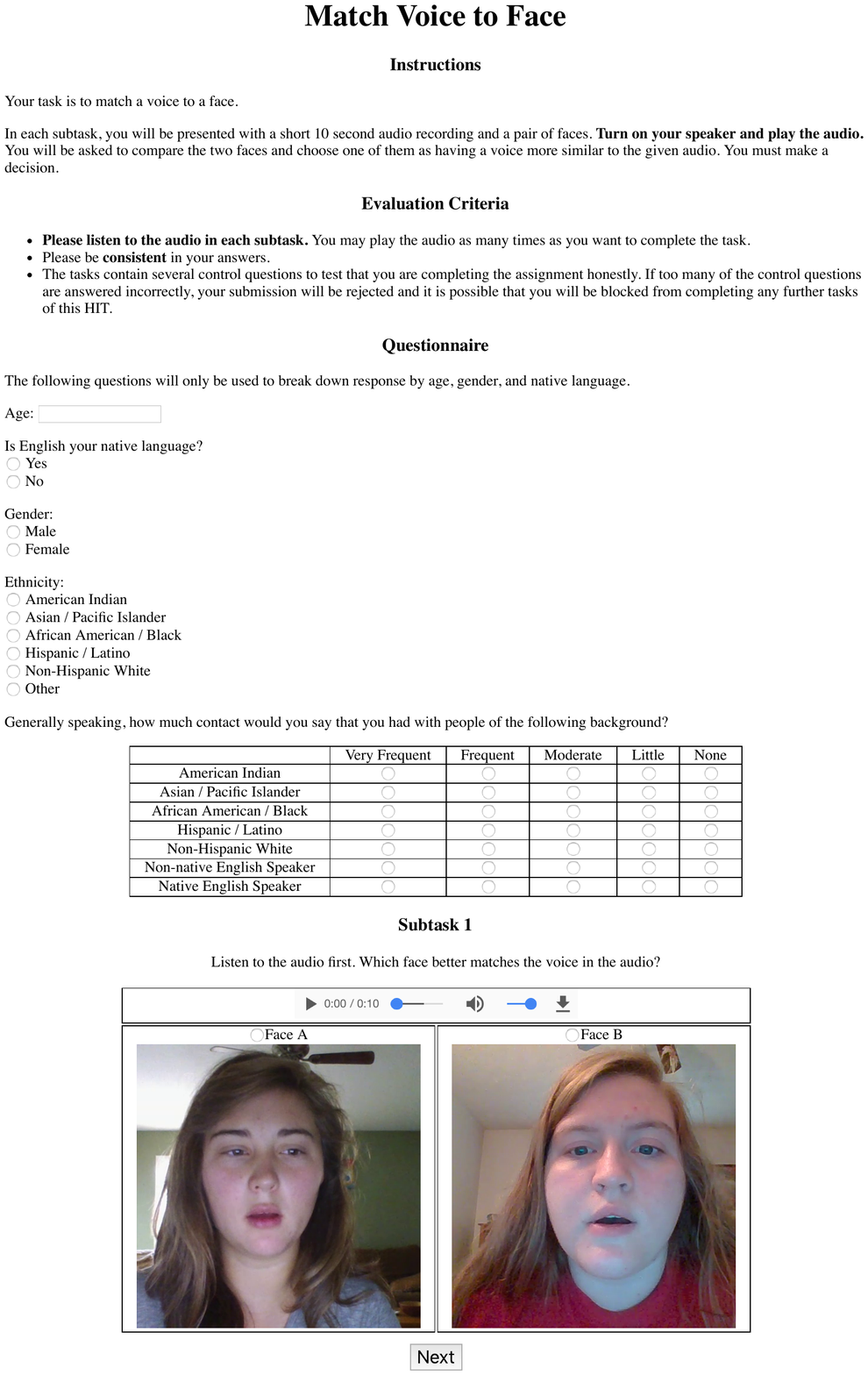}
  \caption{Screenshot of our user study questionnaire used for Experiments 1--3 \mbox{(V $\rightarrow$ F)}. The answers were collected through the web interface of Amazon Mechanical Turk.}
  \label{fig:userstudy_screenshot_v2f}
\end{figure}

\begin{figure}
  \centering
  \includegraphics[width=\linewidth,trim={2cm 1.5cm 2cm 1cm},clip]{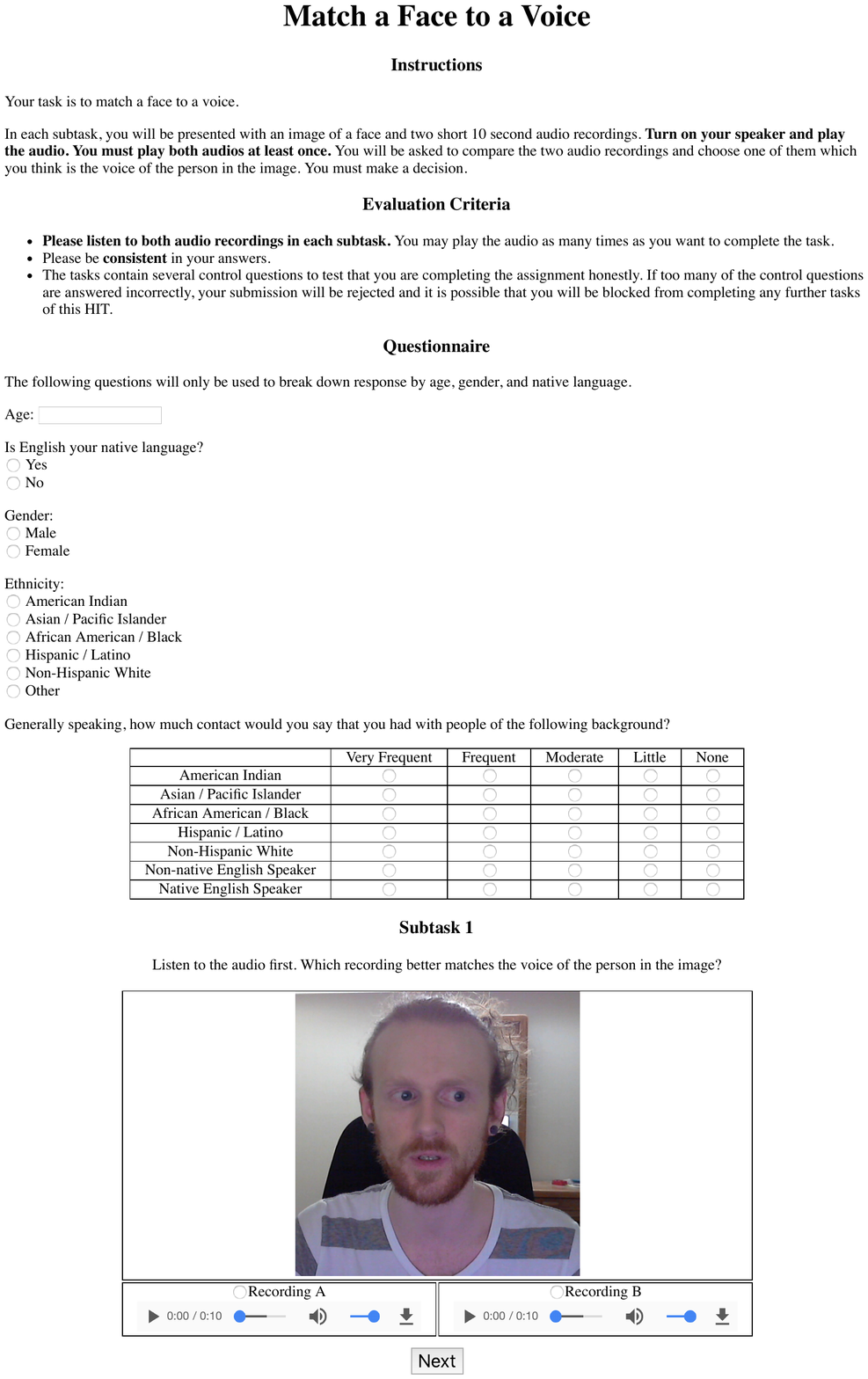}
  \caption{Screenshot of our user study questionnaire used for Experiments 4 \mbox{(F $\rightarrow$ V)}. The answers were collected through the web interface of Amazon Mechanical Turk.}
  \label{fig:userstudy_screenshot_f2v}
\end{figure}

\figreflist{userstudy_screenshot_v2f}{userstudy_screenshot_f2v} show the questionnaire and example subtasks we used for Experiments 1--3 and Experiment 4, respectively, in \secref{userstudy} of the main paper. Actual subtasks are randomized every run.

\subsection{Dataset Acquisition}

\begin{figure}
  \centering
  \includegraphics[width=0.497\linewidth]{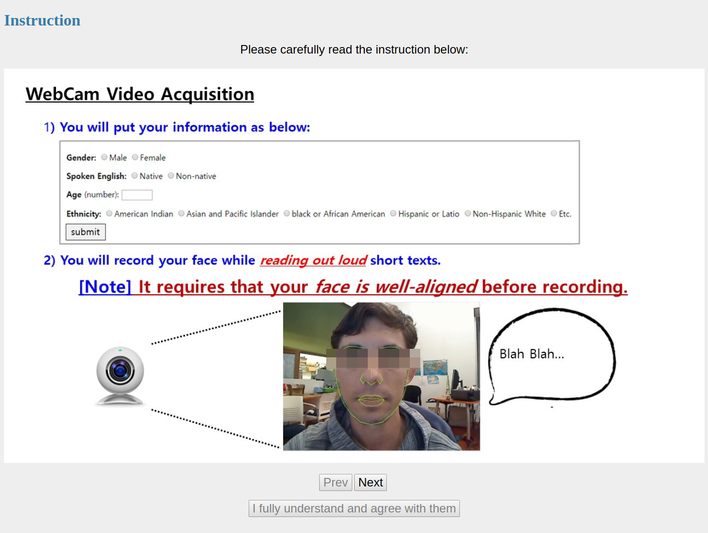}\hfill%
  \includegraphics[width=0.497\linewidth]{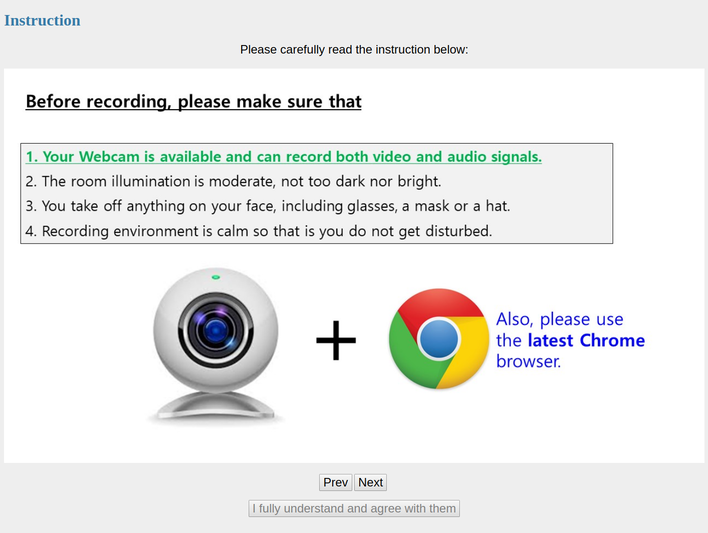}\\[0.05cm]
  \includegraphics[width=0.497\linewidth]{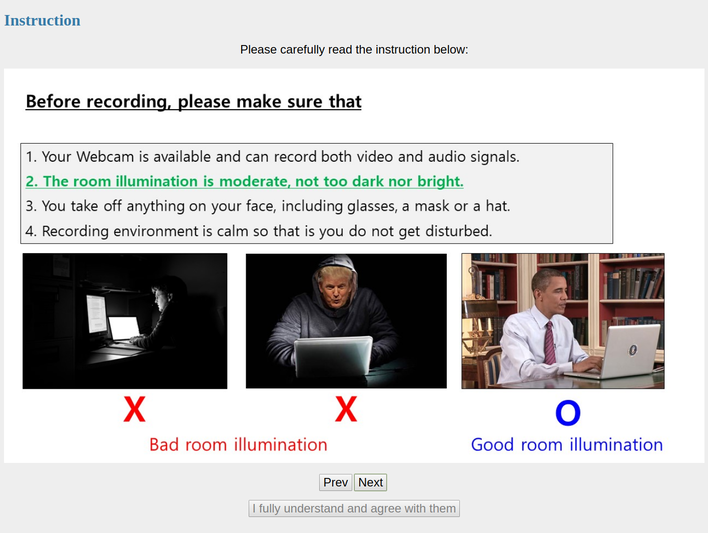}\hfill%
  \includegraphics[width=0.497\linewidth]{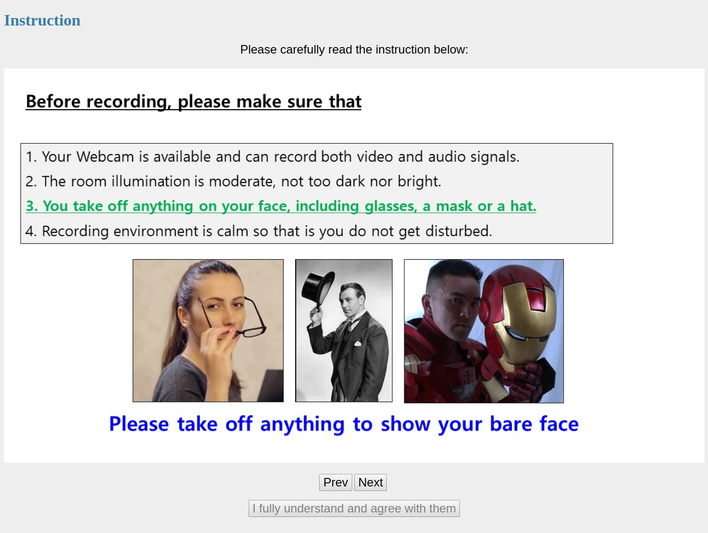}\\[0.05cm]
  \includegraphics[width=0.497\linewidth]{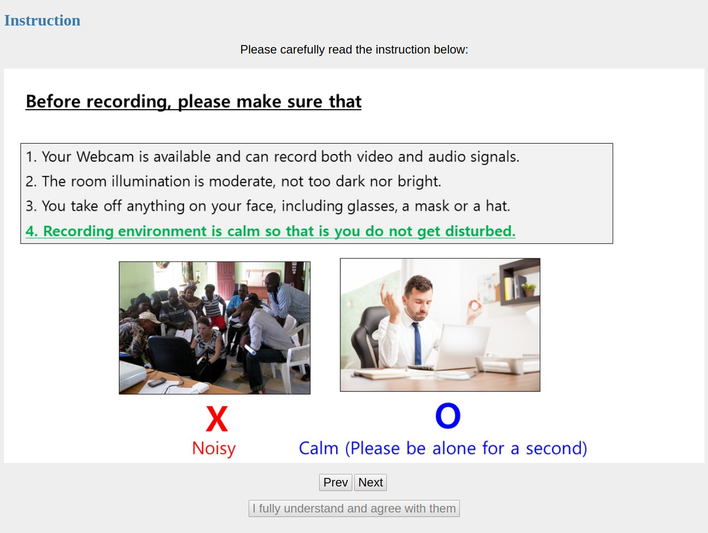}\hfill%
  \includegraphics[width=0.497\linewidth]{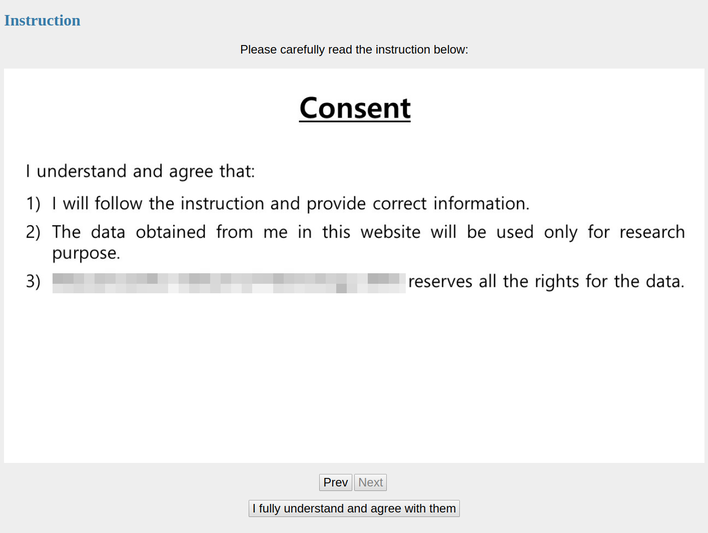}\\
  \caption{Screenshots of the instructions used for collecting our dataset for user studies. The web application was deployed through Amazon Mechanical Turk. Part of screenshots are masked out for anonymity.}
  \label{fig:datacollection_instructions_screenshots}
\end{figure}

\begin{figure}
  \centering
  \includegraphics[width=0.497\linewidth]{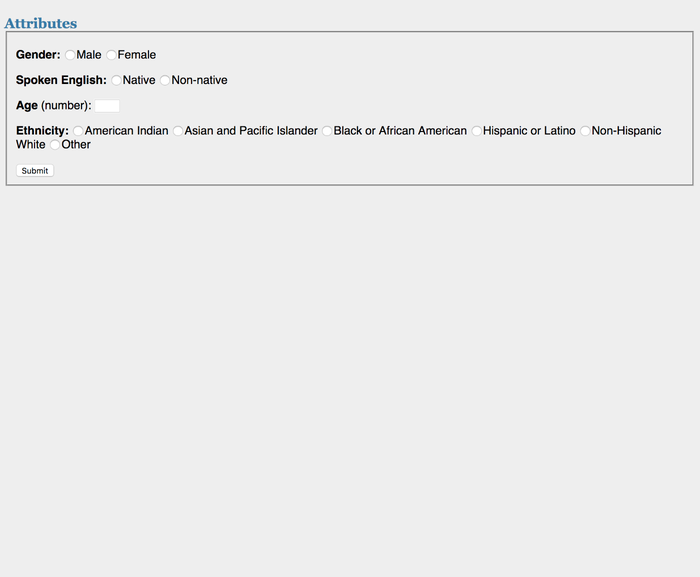}\hfill%
  \includegraphics[width=0.497\linewidth]{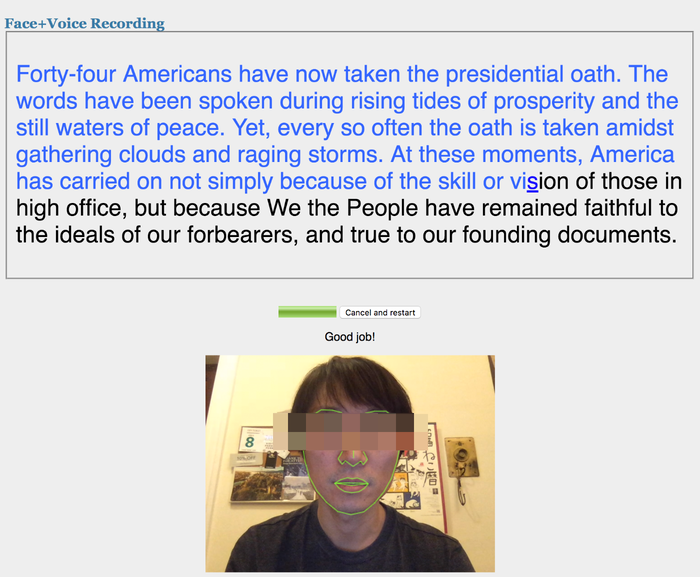}\\
  \caption{Screenshots of a recording session of our dataset for user studies. The web application was deployed through Amazon Mechanical Turk. Part of screenshots are masked out for anonymity.}
  \label{fig:datacollection_recording_screenshots}
\end{figure}

\begin{figure}[t]
  \centering
  \includegraphics[width=0.123\linewidth]{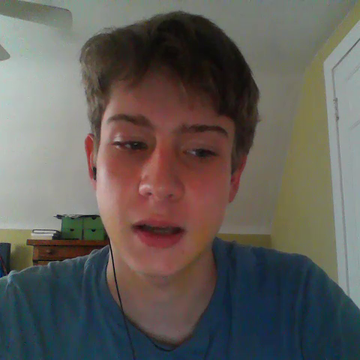}\hfill%
  \includegraphics[width=0.123\linewidth]{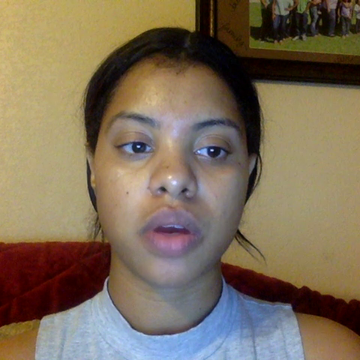}\hfill%
  \includegraphics[width=0.123\linewidth]{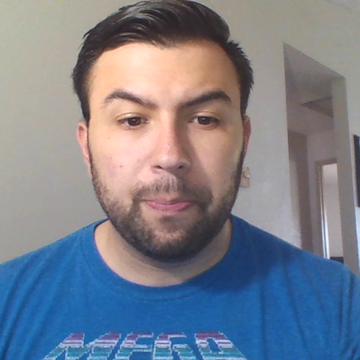}\hfill%
  \includegraphics[width=0.123\linewidth]{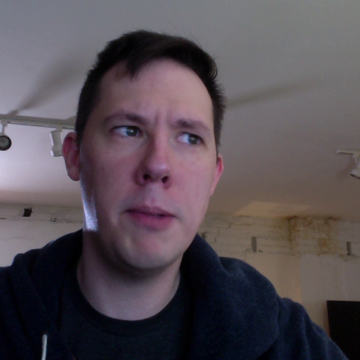}\hfill%
  \includegraphics[width=0.123\linewidth]{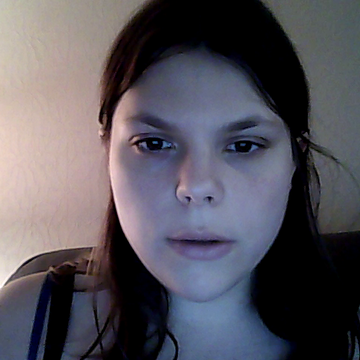}\hfill%
  \includegraphics[width=0.123\linewidth]{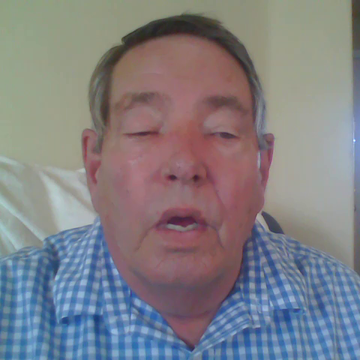}\hfill%
  \includegraphics[width=0.123\linewidth]{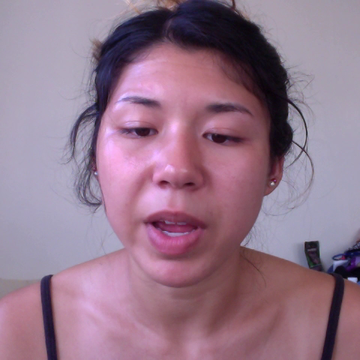}\hfill%
  \includegraphics[width=0.123\linewidth]{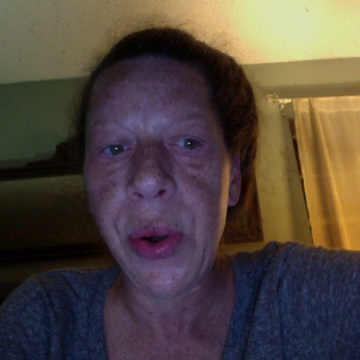}\\
  \includegraphics[width=0.123\linewidth]{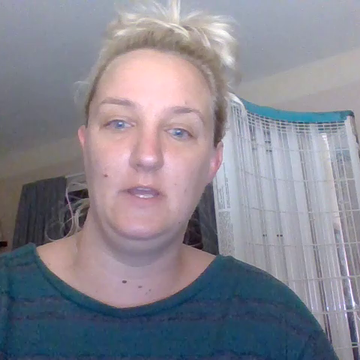}\hfill%
  \includegraphics[width=0.123\linewidth]{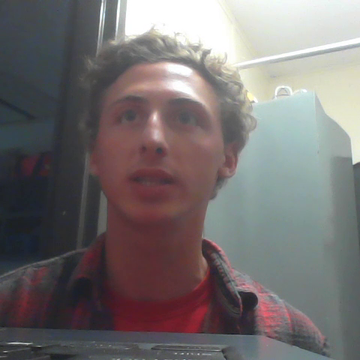}\hfill%
  \includegraphics[width=0.123\linewidth]{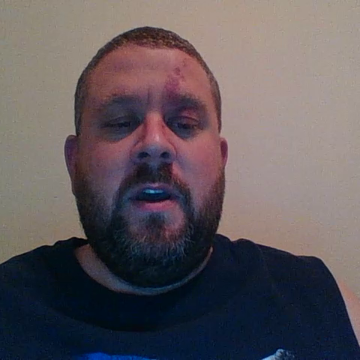}\hfill%
  \includegraphics[width=0.123\linewidth]{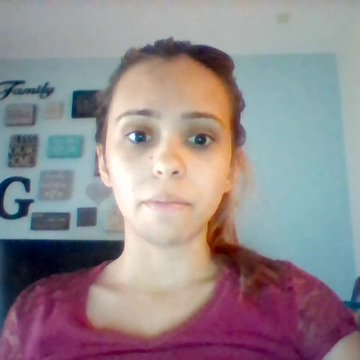}\hfill%
  \includegraphics[width=0.123\linewidth]{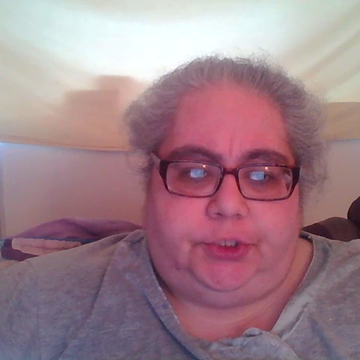}\hfill%
  \includegraphics[width=0.123\linewidth]{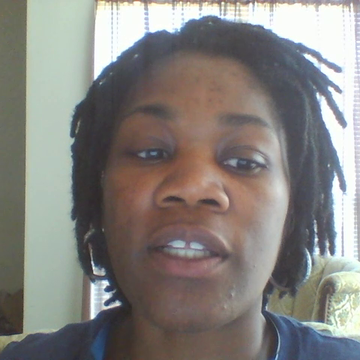}\hfill%
  \includegraphics[width=0.123\linewidth]{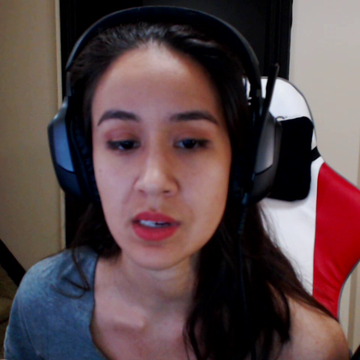}\hfill%
  \includegraphics[width=0.123\linewidth]{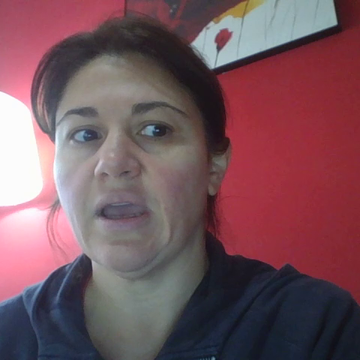}\\
  \includegraphics[width=0.123\linewidth]{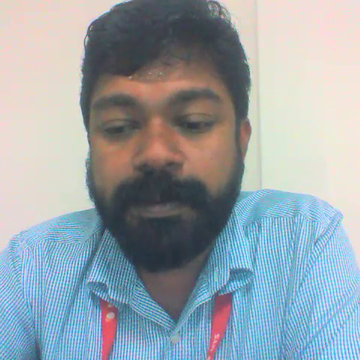}\hfill%
  \includegraphics[width=0.123\linewidth]{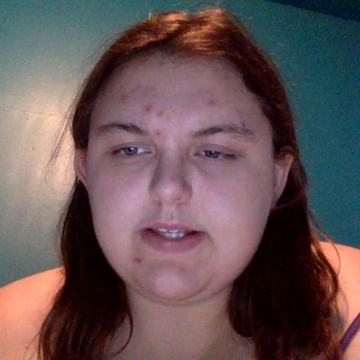}\hfill%
  \includegraphics[width=0.123\linewidth]{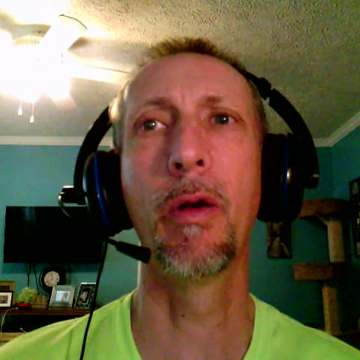}\hfill%
  \includegraphics[width=0.123\linewidth]{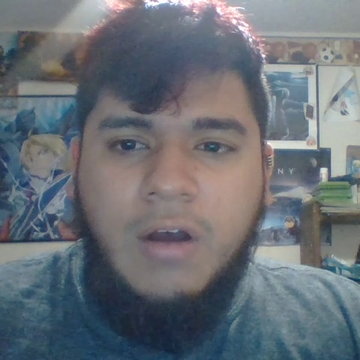}\hfill%
  \includegraphics[width=0.123\linewidth]{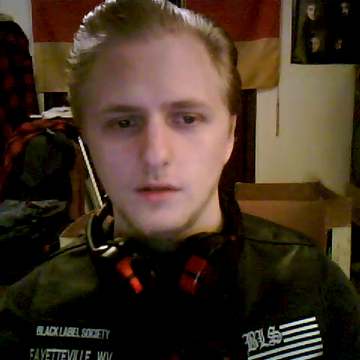}\hfill%
  \includegraphics[width=0.123\linewidth]{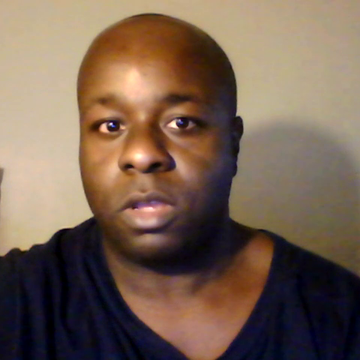}\hfill%
  \includegraphics[width=0.123\linewidth]{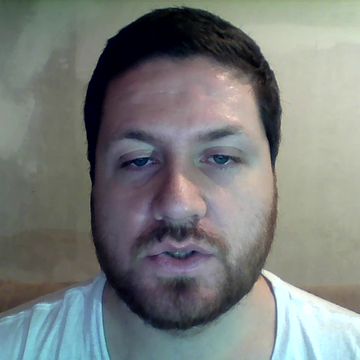}\hfill%
  \includegraphics[width=0.123\linewidth]{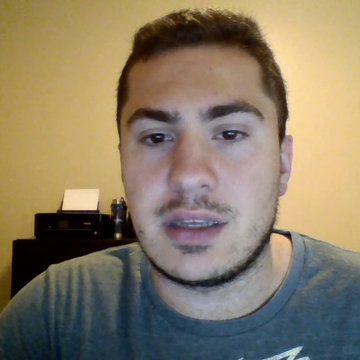}
  \caption{A few face samples from our collected dataset. See the accompanied video for voice playback.}
  \label{fig:amt_faces}
\end{figure}

\figref{datacollection_instructions_screenshots} shows the instructions for data collection. Every participant was requested to read the instructions carefully and to consent to the use of the collected dataset for research purposes. \figref{datacollection_recording_screenshots} shows the questionnaire for demographic information and an example recording session.
In order to encourage constant reading speed, words are sequentially highlighted in the script, similar to popular karaoke interfaces. Furthermore, to normalize the head position, we provide facial markers where participants can align their face to a centered front-facing position. Feedback about the alignment is provided using the \texttt{clmtrackr} library, a JavaScript implementation of the face tracking model of Saragih et al.~\cite{Saragih2009}.
Participants can repeat the recording session until they are satisfied.
From the collected video recordings of them speaking, we extract still face images and ten-second-long audio clips containing their voices.
We manually cleaned the collected data, for example, removing recordings with loud background noise or low audio/video quality.
A few example face images are shown in \figref{amt_faces}; for audio playback, browse our dataset at \url{http://facevoice.csail.mit.edu}.
The text is chosen from the following pool:
\begin{itemize}
  \item ``Forty-four Americans have now taken the presidential oath. The words have been spoken during rising tides of prosperity and the still waters of peace. Yet, every so often the oath is taken amidst gathering clouds and raging storms. At these moments, America has carried on not simply because of the skill or vision of those in high office, but because We the People have remained faithful to the ideals of our forbearers, and true to our founding documents.''
  \item ``That we are in the midst of crisis is now well understood. Our nation is at war, against a far-reaching network of violence and hatred. Our economy is badly weakened, a consequence of greed and irresponsibility on the part of some, but also our collective failure to make hard choices and prepare the nation for a new age. Homes have been lost; jobs shed; businesses shuttered. Our health care is too costly; our schools fail too many; and each day brings further evidence that the ways we use energy strengthen our adversaries and threaten our planet.''
  \item ``In reaffirming the greatness of our nation, we understand that greatness is never a given. It must be earned. Our journey has never been one of short-cuts or settling for less. It has not been the path for the faint-hearted - for those who prefer leisure over work, or seek only the pleasures of riches and fame. Rather, it has been the path for the risk-takers, the doers, the makers of things - some celebrated but more often men and women obscure in their labor, who have carried us up the long, rugged path towards prosperity and freedom.''
  \item ``For us, they fought and died, in places like Concord and Gettysburg; Normandy and Khe Sahn. Time and again these men and women struggled and sacrificed and worked till their hands were raw so that we might live a better life. They saw America as bigger than the sum of our individual ambitions; greater than all the differences of birth or wealth or faction.''
  \item ``To the Muslim world, we seek a new way forward, based on mutual interest and mutual respect. To those leaders around the globe who seek to sow conflict, or blame their society's ills on the West - know that your people will judge you on what you can build, not what you destroy. To those who cling to power through corruption and deceit and the silencing of dissent, know that you are on the wrong side of history; but that we will extend a hand if you are willing to unclench your fist.''
  \item ``For as much as government can do and must do, it is ultimately the faith and determination of the American people upon which this nation relies. It is the kindness to take in a stranger when the levees break, the selflessness of workers who would rather cut their hours than see a friend lose their job which sees us through our darkest hours. It is the firefighter's courage to storm a stairway filled with smoke, but also a parent's willingness to nurture a child, that finally decides our fate.''
  \item ``So let us mark this day with remembrance, of who we are and how far we have traveled. In the year of America's birth, in the coldest of months, a small band of patriots huddled by dying campfires on the shores of an icy river. The capital was abandoned. The enemy was advancing. The snow was stained with blood. At a moment when the outcome of our revolution was most in doubt, the father of our nation ordered these words be read to the people:''
  \item ``America. In the face of our common dangers, in this winter of our hardship, let us remember these timeless words. With hope and virtue, let us brave once more the icy currents, and endure what storms may come. Let it be said by our children's children that when we were tested we refused to let this journey end, that we did not turn back nor did we falter; and with eyes fixed on the horizon and God's grace upon us, we carried forth that great gift of freedom and delivered it safely to future generations.''
\end{itemize}

\section{Evaluations on Machine Performance}

In this section, further evaluations and visualizations of our learned representations omitted from \secref{visualization} of the main paper are provided. We conclude this section with additional discussions.

\subsection{Further Evaluations on the Learned Representation}

\paragraph{The t-SNE visualizations.}

\begin{figure}[!p]
  \centering
  \includegraphics[width=0.495\linewidth]{figures/tsne/vctest_s1000_f2v_voice_p50/tsne_gender}\hfill%
  \includegraphics[width=0.495\linewidth]{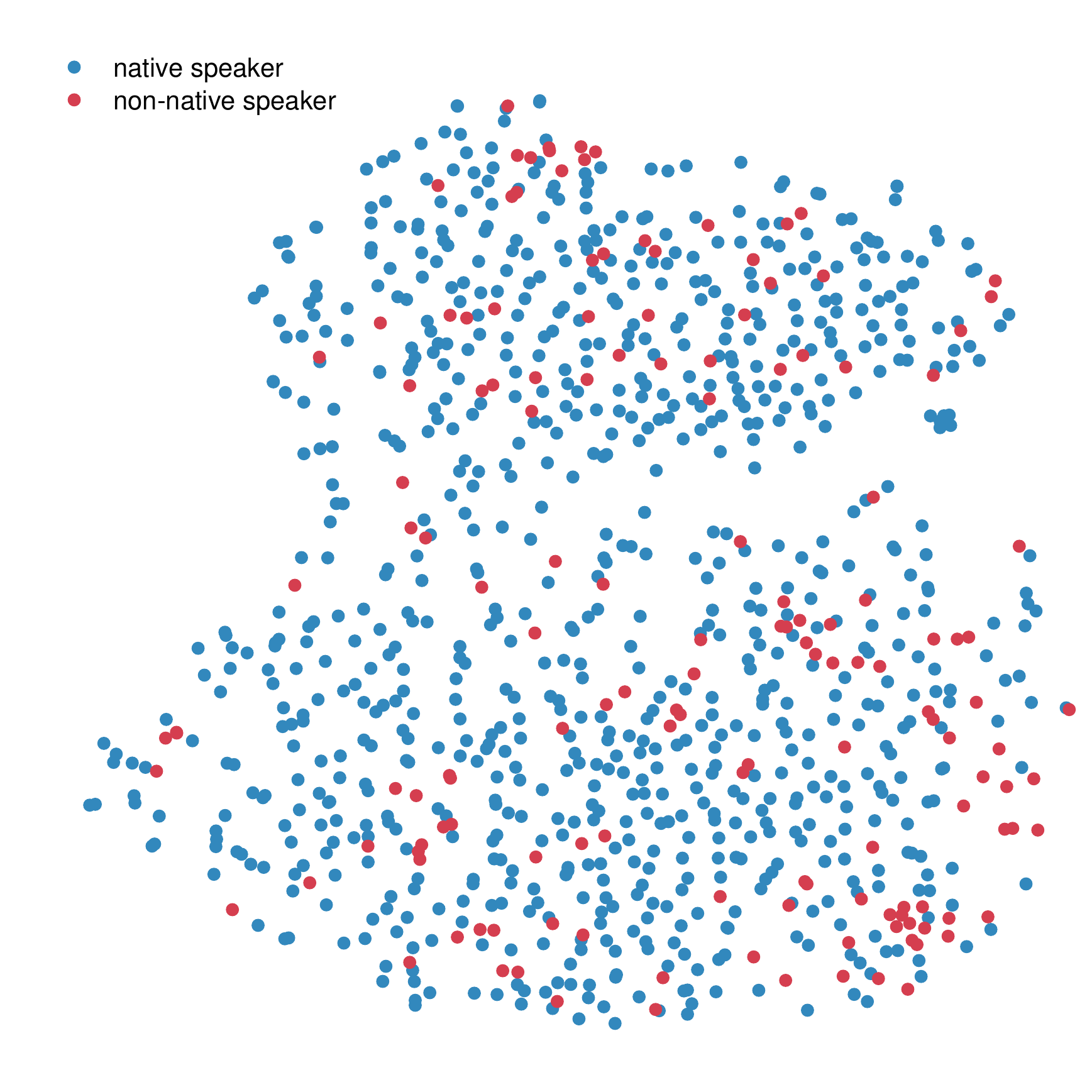}\\
  \parbox{0.495\linewidth}{\centering\scriptsize{}(a) Voice; gender}\hfill%
  \parbox{0.495\linewidth}{\centering\scriptsize{}(b) Voice; first language}\\
  \includegraphics[width=0.495\linewidth]{figures/tsne/vctest_s1000_f2v_voice_p50/tsne_age}\hfill%
  \includegraphics[width=0.495\linewidth]{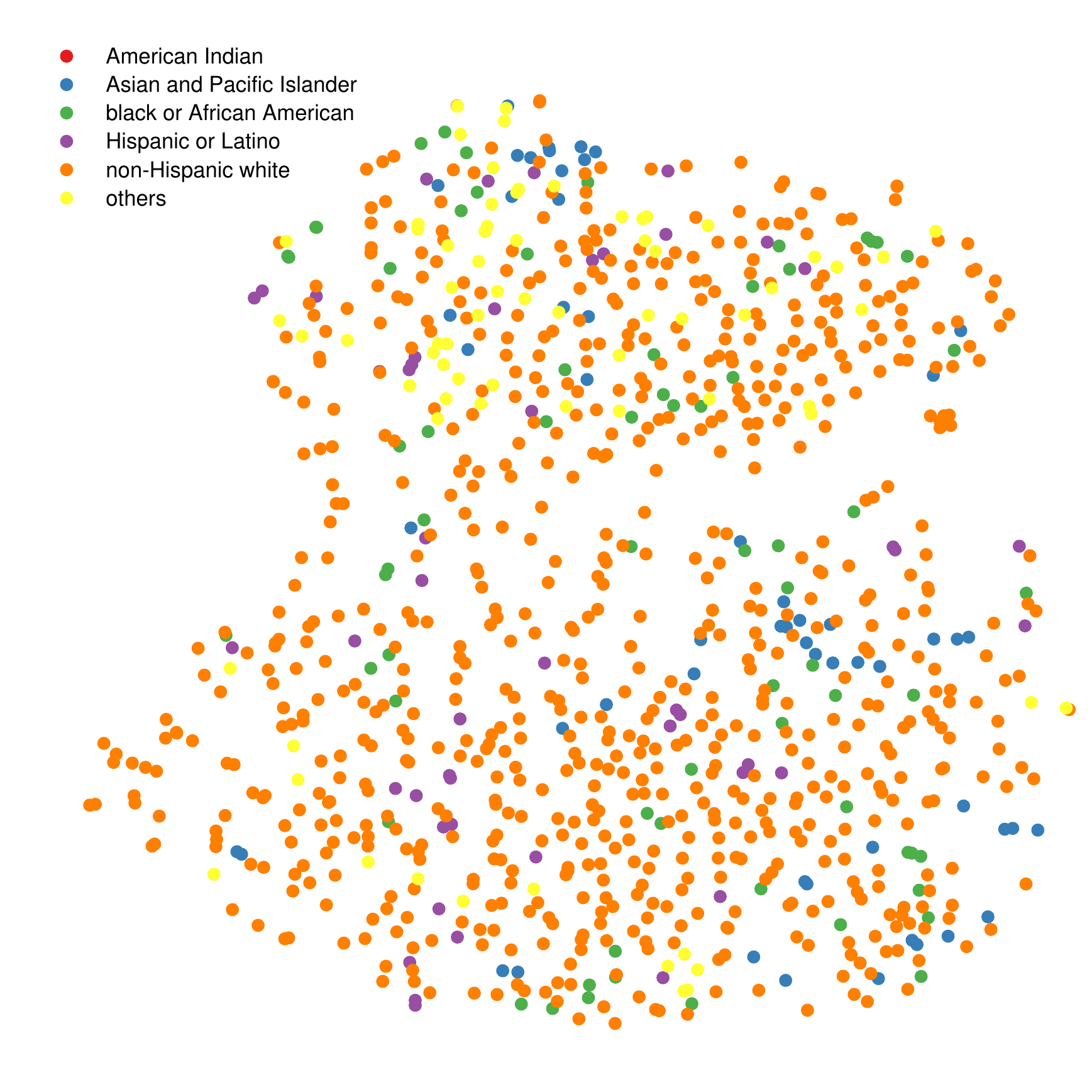}\\
  \parbox{0.495\linewidth}{\centering\scriptsize{}(c) Voice; age}\hfill%
  \parbox{0.495\linewidth}{\centering\scriptsize{}(d) Voice; ethnicity}\\
  \caption{The t-SNE visualization of the \emph{voice} representations of VoxCeleb test samples. 1,000 random samples are drawn from the test set and shown with four demographic attributes. (c) The color code depicts continuous values, while the legend shows only the minimum and the maximum values; the rest encode categorical values. See the main paper for the t-SNE with the voice identity marked.}
  \label{fig:tsne-voice}
\end{figure}

\figreflist{tsne-voice}{tsne-face} show the t-SNE visualization~\cite{VanDerMaaten2008} of our learned voice and face representation, respectively.
We drew 1,000 random samples and used our annotations to color-code the sample points according to their four demographic attributes. See \figref{tsne} of the main paper for the t-SNE visualized with face/voice identities.
Note that our network has not seen any of the demographic attributes during training.

\begin{figure}[!p]
  \centering
  \includegraphics[width=0.495\linewidth]{figures/tsne/vctest_s1000_v2f_face_p50/tsne_gender}\hfill%
  \includegraphics[width=0.495\linewidth]{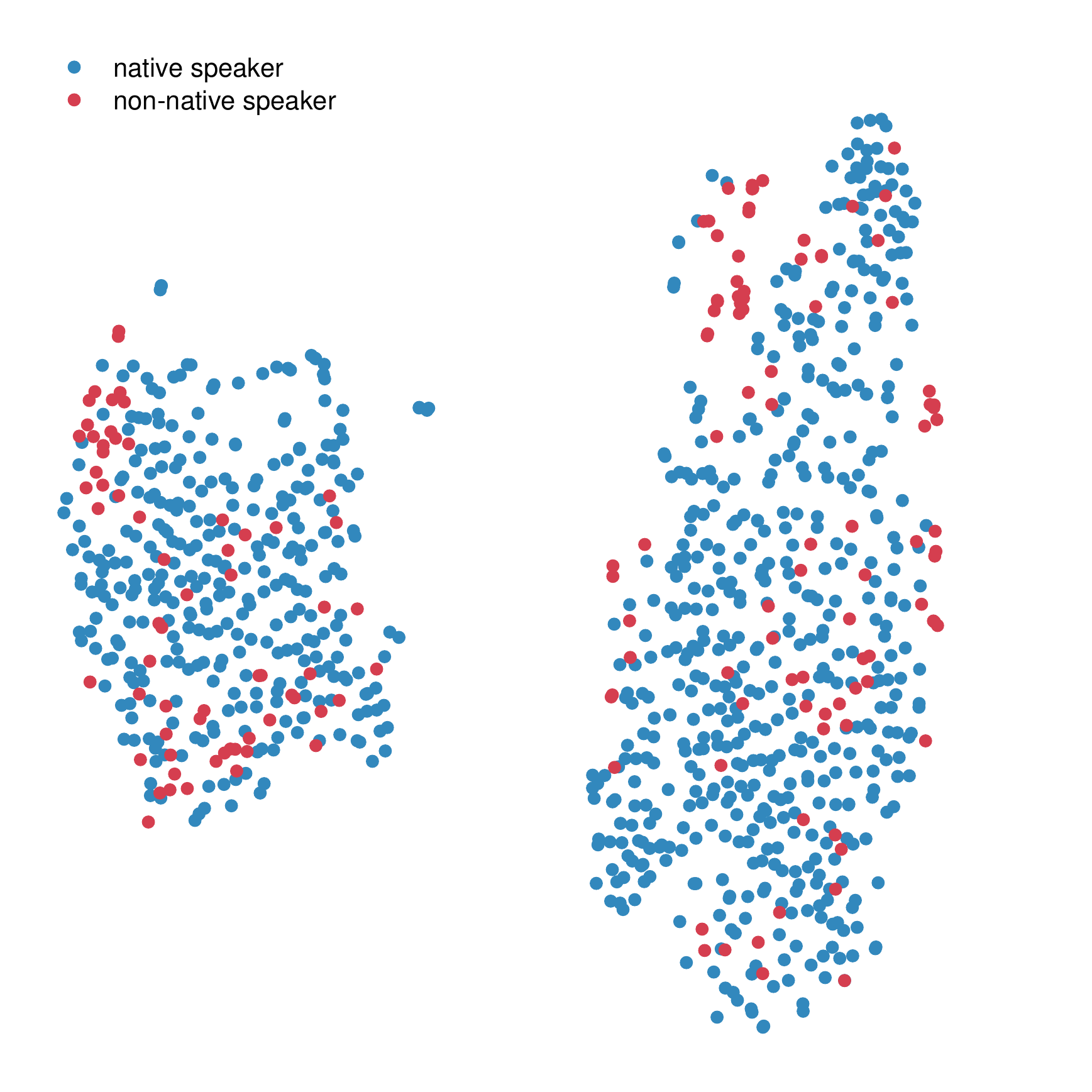}\\
  \parbox{0.495\linewidth}{\centering\scriptsize{}(a) Face; gender}\hfill%
  \parbox{0.495\linewidth}{\centering\scriptsize{}(b) Face; first language}\\
  \includegraphics[width=0.495\linewidth]{figures/tsne/vctest_s1000_v2f_face_p50/tsne_age}\hfill%
  \includegraphics[width=0.495\linewidth]{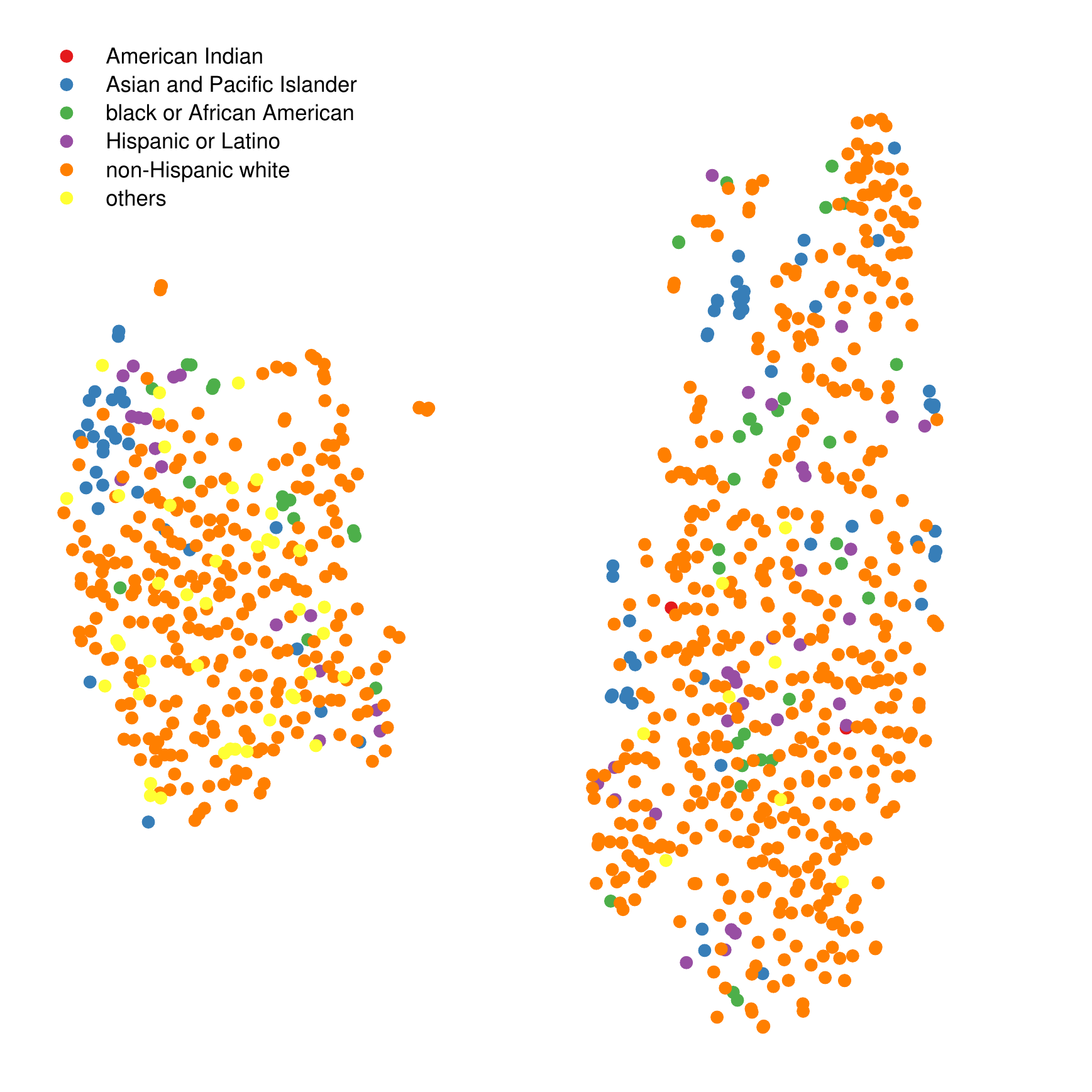}\\
  \parbox{0.495\linewidth}{\centering\scriptsize{}(c) Face; age}\hfill%
  \parbox{0.495\linewidth}{\centering\scriptsize{}(d) Face; ethnicity}\\
  \caption{The t-SNE visualization of the \emph{face} representations of VoxCeleb test samples. 1,000 random samples are drawn from the test set and shown with four demographic attributes. (c) The color code depicts continuous values, while the legend shows only the minimum and the maximum values; the rest encode categorical values. See the main paper for the t-SNE with the face identity marked.}
  \label{fig:tsne-face}
\end{figure}

As discussed in the main paper, the learned representation forms the clearest clusters regarding gender (\figreflistsub{tsne-voice}{tsne-face}{a}), which explains the performance drop when the samples are constrained by gender. While correlated with gender, age shows a distinct grouping from gender (\figreflistsub{tsne-voice}{tsne-face}{c}). In particular in face representations, \figref{tsne-face}c shows the age distributed orthogonal to gender: it increases from bottom to top while the gender is split horizontally.
The t-SNE visualization does not reveal such strong clustering regarding the first language or the ethnic group (\figreflistsub{tsne-voice}{tsne-face}{bd}), and presents only small clusters scattered across the projection. 
As also noted in the main paper, such absence of clustering does not rule out the existence of additional information encoded in our learned representations, which we argue with additional evidences in the following.

\paragraph{More evaluations of linear classifiers on our representations.}

\tabref{attributel-ext} summarizes the quality of linear classifiers for demographic attributes on our learned representations, similar to \tabref{attributel} of the main paper, to demonstrate what information our representation encodes.

\begin{table}[b!]
  \caption{The analysis of encoded information in face and voice representations. This experiment is similar to \tabref{attributel} in the main paper, but with the triplet network trained with a face anchor subnetwork and positive and negative voice subnetworks. (In the main paper, we report the performance with the triplet network trained with a voice anchor subnetwork and positive and negative face subnetworks.) As a probe task, we use the attribute classification task. We report the mean average precision (mAP) with 99$\%$ confidence intervals (CI) obtained from 20 trials of holdout cross validations. We mark the values having confidence intervals that overlap with a random chance with a 5$\%$ margin, i.e., $50\pm5\%$, in red. In cases where the performance is less than or equal to random chance, it is suggested that the representation is not distinctive enough for the classification task. Note that during representation learning, no attribute information was seen by the network.}
  \resizebox{\linewidth}{!}{%
    \setlength{\tabcolsep}{1.5mm}%
    \begin{tabular}{@{}ccccccccccccccc@{}}%
      \toprule
      \multicolumn{2}{@{}l}{\multirow{2}{*}[-.5mm]{Modality}} & \multirow{2}{*}[-.5mm]{Gender} & \multirow{2}{*}[-.5mm]{Fluency} & \multicolumn{5}{c}{Age} & \multicolumn{6}{c}{Ethnicity} \\
      \cmidrule(lr){5-9} \cmidrule(ll){10-15}
      & & & & $<$30 & 30s & 40s & 50s & $\geq$60 & 1 & 2 & 3 & 4 & 5 & 6\\
      \midrule
      \multirow{2}{*}{Face repr.} & mAP & 98.7 & 67.9 & 71.9 & 68.4 & \hlck{57.6} & 63.1 & 81.4 & 90.3 & 79.6 & 81.9 & 71.1 & 67.8 & 74.3 \\
      & CI & ${\pm}$2.6 & ${\pm}$4.1 & ${\pm}$9.9 & ${\pm}$3.7 & \hlck{$\pm$3.8} & ${\pm}$7 & ${\pm}$3.8 & ${\pm}$4.3 & ${\pm}$6 & ${\pm}$5.5 & ${\pm}$5.5 & ${\pm}$6.9 & ${\pm}$5.5 \\
      \cmidrule{1-15}
      \multirow{2}{*}{Voice repr.} & mAP & 93.1 & \hlck{58.2} & 65.7 & \hlck{56.7} & \hlck{52.7} & \hlck{56.7} & 62.9 & 94.5 & 71.1 & \hlck{58.4} & 61.1 & \hlck{55.8} & 68.5 \\
      & CI & ${\pm}$1.8 & ${\pm}$3.4 & ${\pm}$3.9 & ${\pm}$3.6 & ${\pm}$1.4 & ${\pm}$3.7 & ${\pm}$5.9 & ${\pm}$2 & ${\pm}$7.1 & ${\pm}$4.6 & ${\pm}$2.7 & ${\pm}$4.6 & ${\pm}$9.3 \\
      \bottomrule
    \end{tabular}%
  }%
  \label{tab:attributel-ext}
\end{table}

As evidenced by a high classification precision, the representation provides the most distinctive information for gender classification, which is consistent with the distributions observed in \figreflistsub{tsne-voice}{tsne-face}{a}.
Age is a continuous attribute as demonstrated in \figreflistsub{tsne-voice}{tsne-face}{a}, and grouping into a discrete set of ranges (as in \tabref{attributel}) makes the classification results more conservative: i.e., \figref{tsne-face}f shows overall smooth transition in age, but far from perfect ordering especially in mid-ages, resulting in less decisive age classification results shown in \tabref{attributel-ext}.
We note that the analyses of \tabref{attributel-ext} and \figreflist{tsne-voice}{tsne-face} (as well as \tabref{attributel} and \figref{tsne} in the main paper) are complementary to, and consistent with, each other.
t-SNE is an unsupervised method for visualization which typically reveals dominant information encoded in the representation, while
the experiment in \tabref{attributel-ext} (and \tabref{attributel} of the main paper) exploits supervised information to reveal hidden information in the representation.

\subsection{Further Discussions}

\paragraph{Comparisons to binary classification.}

We experimented with a classification network inspired by the ``$L^3$ network''~\cite{Arandjelovic2017}, an audiovisual correlation classifier, and Nagrani et al.'s model~\cite{Nagrani2018}, and trained to do binary classification: given a face and a voice, whether or not the two belong to the same identity. Here we detail the construction of the network. The classification network shares the same subnetworks as our architecture based on the triplet loss, but the two 512-d feature vectors average-pooled from the \texttt{conf5\_3} and \texttt{conv6} of VGG16 and SoundNet, respectively, are concatenated to form a 1024-d vector, which is then fed to two 128-d fully-connected layers, in succession, followed by a 2-d fully-connected layer and the softmax activation. The class probability of the positive association is used as a score to measure the similarity of the face and the voice, hence for gauging the distances between a given voice (face) and two candidate faces (voices). The candidate with the higher similarity score is taken as the matching pair.

\paragraph{Dimensions of fully connected layers.}

\begin{table}[t]
  \centering
  \caption{Test accuracy with varying fully connected layer dimensions (and thus our representation dimensions). For smaller dimensions, the last convolutional layer of each subnetwork is average-pooled globally before fed to the first fully-connected layer; for the dimensions larger than the filter dimension of the last convolutional layer (512-d), it was average-pooled with the factor of 2 along each non-singleton spatial dimension.}
  \begin{tabular}{@{}%
      L{0.16\linewidth}%
      L{0.03\linewidth}%
      L{0.14\linewidth}%
      L{0.14\linewidth}%
      L{0.14\linewidth}%
      L{0.14\linewidth}%
      L{0.14\linewidth}%
      @{}}
    \toprule
    \multirow{2}{*}{Experiments} && \multicolumn{2}{l}{Global spatial pooling} & \multicolumn{3}{l}{2$\times$ spatial pooling} \\
    \cmidrule(r){3-4} \cmidrule{5-7}
                                 && 128-d  & 512-d  & 1024-d & 2048-d & 4096-d \\
    \midrule
    --                           && 78.2\% & 77.4\% & 77.9\% & 77.7\% & 77.6\% \\
    G/E/F/A                      && 59.0\% & 57.6\% & 58.5\% & 58.1\% & 58.2\% \\
    \bottomrule
  \end{tabular}
  \label{tab:fc_layer_test}
\end{table}

We measured the test accuracy with varying dimensions of the fully connected layers (and thus the representation vectors), which is tabulated in \tabref{fc_layer_test}. While this did not have a significant influence on test accuracy, generally, narrower fully connected layers resulted in slightly better performance.

\paragraph{Further details on training.}

The batch size and the learning rate were chosen by grid search within the machine limit. Decaying learning rates and mining hard negative samples helped stabilize training and prevent from overfitting to training data, but did not contribute much to improve accuracy. The timing and amount of decaying were set empirically.
A standard data augmentation scheme was used optionally: face images are randomly cropped around the face region by $-40\%$ to $20\%$, rotated for a random angle between $\pm 15\degree$, and horizontally flipped randomly. Negative cropping means including more background. Image brightness and contrast as well as audio volume are jittered up to $\pm 20\%$. We trained our network both with and without data augmentation under the same setup outlined below, but did not find significant difference in performance.
This could be due to the large sample size (over 100k) and great diversity of the VoxCeleb dataset we used for training.
Our model was implemented using TensorFlow and trained on an NVIDIA Titan X (Pascal) with 12\,Gb RAM. Training typically takes less than a day on a single GPU.